\documentclass[10pt,twocolumn,letterpaper]{article}
\usepackage{titling}

\usepackage{iccv}
\usepackage{times}
\usepackage{epsfig}
\usepackage{graphicx}
\usepackage{amsmath}
\usepackage{amssymb}
\usepackage{amsthm}

\usepackage[]{graphicx}
\usepackage{amsfonts, amssymb}
\usepackage{array}
\usepackage{mdwmath}
\usepackage{mdwtab}
\usepackage[tight,footnotesize]{subfigure}
\usepackage{algorithm}	
\usepackage{algpseudocode}
\usepackage{balance}
\usepackage{comment}
\usepackage{gensymb}
\usepackage{multirow}
\usepackage{comment}
\usepackage[]{appendix}
\usepackage{bm}
\usepackage{mathrsfs}
\usepackage[utf8]{inputenc}
\usepackage[english]{babel}
\usepackage[bottom]{footmisc}
\usepackage{mathrsfs}

\newcommand{\PreserveBackslash}[1]{\let\temp=\\#1\let\\=\temp}
\newcolumntype{C}[1]{>{\PreserveBackslash\centering}p{#1}}
\newcolumntype{R}[1]{>{\PreserveBackslash\raggedleft}p{#1}}
\newcolumntype{L}[1]{>{\PreserveBackslash\raggedright}p{#1}}

\usepackage[pagebackref=true,breaklinks=true,letterpaper=true,colorlinks,bookmarks=false]{hyperref}

\iccvfinalcopy 

\def\iccvPaperID{4168} 

\ificcvfinal\fi

\title{MINE: Towards Continuous Depth MPI with NeRF for Novel View Synthesis}

\ificcvfinal
\usepackage{authblk}

\makeatletter
\renewcommand\AB@affilsepx{, \protect\Affilfont}
\makeatother
\author[1*]{Jiaxin Li}
\author[1*]{Zijian Feng}
\author[1]{Qi She}
\author[1]{Henghui Ding}
\author[1]{Changhu Wang}
\author[2]{Gim Hee Lee}
\affil[1]{ByteDance}
\affil[2]{National University of Singapore}
\fi
\begin{document}
\maketitle

\begin{abstract}
In this paper, we propose \textbf{MINE} to perform novel view synthesis and depth estimation via dense 3D reconstruction from a single image. Our approach is a continuous depth generalization of the \textbf{M}ultiplane \textbf{I}mages (MPI) by introducing the \textbf{NE}ural radiance fields (NeRF). Given a single image as input, MINE predicts a 4-channel image (RGB and volume density) at arbitrary depth values to jointly reconstruct the camera frustum and fill in occluded contents. The reconstructed and inpainted frustum can then be easily rendered into novel RGB or depth views using differentiable rendering. Extensive experiments on RealEstate10K, KITTI and Flowers Light Fields show that our MINE outperforms state-of-the-art by a large margin in novel view synthesis. We also achieve competitive results in depth estimation on iBims-1 and NYU-v2 without annotated depth supervision. Our source code is available at \small \url{https://github.com/vincentfung13/MINE}.

\end{abstract} 
\ificcvfinal
\renewcommand*{\thefootnote}{\fnsymbol{footnote}}
\footnotetext[1]{Equal contribution.}
\renewcommand*{\thefootnote}{\arabic{footnote}}
\fi
\section{Introduction} \label{sec_intro}

Interactive 3D scene is a fascinating way to achieve immersive user experience similar to augmented/virtual reality. To automate or simplify the creation of 3D scenes, increasing efforts are invested on novel view synthesis from a single or multiple image(s) that enables rendering at arbitrary camera poses according to user's interaction. Despite its usefulness, the novel view synthesis problem is challenging because it requires precise geometry understanding, and inpainting of the occluded geometry and textures.

To tackle the problem of view synthesis, most existing methods focus on the design of 3D or 2.5D representations of the scene, and the rendering techniques of novel views. A straightforward idea is to perform Structure-from-Motion (SfM)~\cite{schoenberger2016mvs,schoenberger2016sfm} or monocular/multiview depth estimations~\cite{monodepth2,monodepth17,zhou2017unsupervised,sfmnet2017} to recover the 3D scene. Unfortunately, this naive approach is insufficient to  acquire accurate dense 3D geometry and fill in the occluded contents of the scene. Consequently, this results to distorsion and artifacts in the rendered novel views. To alleviate this problem, more sophisticated representations including Layed Depth Image (LDI) \cite{lsiTulsiani18,shade1998layered}, Multiplane Images (MPI) \cite{Tucker_2020_CVPR} are used with deep networks to recovered 2.5D information from single / multiple images. However, 2.5D approaches usually suffer from limited resolution to represent the full 3D scene. 

Recently, the MPI \cite{Tucker_2020_CVPR} representation attracts a lot of attentions. Specifically, it is a deep network supervised with other image views of the same scene to lift a RGB image into multiple planes of RGB and alpha values. Novel views are then rendered by performing homography warping and integral over the planes. Despite its success, the MPI method fails to represent continuous 3D space effectively. Its depth-wise resolution is limited by the number of discrete planes, and thus the MPIs cannot be converted to other 3D representations such as mesh, point cloud, etc.
In contrast, the Neural Radiance Fields (NeRF) \cite{mildenhall2020nerf} is concurrently proposed to recover 3D information from images using a Multi-layer Perceptron (MLP). The MLP takes a 3D position and a 2D viewing direction as input to predict the RGB and volume occupancy density at that query position. Although NeRF produces high quality 3D structures and novel views, it has to be trained per scene, i.e. one MLP represents only one scene.
\begin{figure}[t!] \centering
\includegraphics[width=0.48\textwidth]{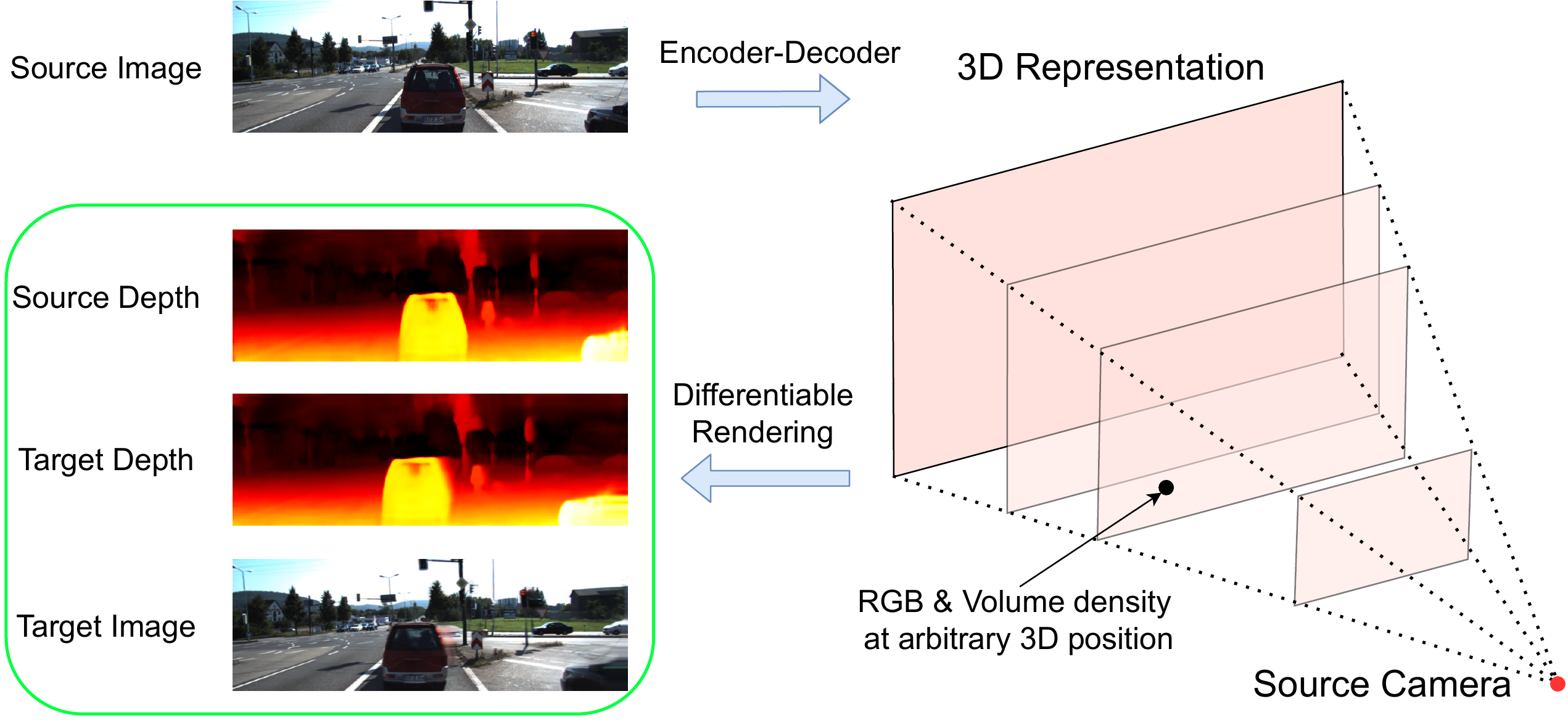}
\caption{Overview of our proposed method.}\label{fig_overall}
\vspace{-12pt}
\end{figure}
%

In view that MPI~\cite{Tucker_2020_CVPR} is unable to represent the full 3D space, 
we propose MINE that generalizes MPI to a continuous 3D representation similar to NeRF~\cite{mildenhall2020nerf}.
Specifically, an input image is first fed into an encoder network to obtain the image features. A decoder network then takes these image features and an arbitrary depth as inputs to produce a 4-channel, i.e. RGB and volume density values, plane fronto-parallel to the input camera. As shown in Sec.~\ref{sec_appr_3drep}, our MINE can effectively reconstructs the camera frustum in full 3D space since the plane depth is arbitrary.
We prove in Sec.~\ref{sec_appr_mpi} that the MPI \cite{Tucker_2020_CVPR} representation is a limited special case of our approach. 
Our main contributions are: \vspace{-1mm}
\begin{itemize}
    \item Performs continuous and occlusion-inpainted 3D reconstruction from a single image. \vspace{-8pt}
    \item Our MINE is a continuous depth generalization of the MPI by introducing the NeRF idea. \vspace{-8pt}
    \item Significantly outperforms existing state-of-the-art methods in indoor and outdoor view synthesis and depth estimation.
\end{itemize}
\section{Related Work}  \label{sec_related_work}

\paragraph{Explicit 3D representations for view synthesis.} Early works on light fields \cite{10.1145/237170.237199,cohen1996the,gortler2001} achieve view synthesis by interplotating nearby views given a set of input images. A recent work \cite{Srinivasan_2017_ICCV}, predicts the entire light field from a single image. Volumetric representations \cite{space_carving_1998, voxel_coloring_1997, szeliski1999stereo, lsmKarHM2017, Henzler_2020_CVPR, sitzmann2019deepvoxels} and view synthesis with predicted depth maps \cite{Niklaus_TOG_2019, Wiles_2020_CVPR} have also been studied intensively. Recently, layered representations, specifically layered depth image (LDI) \cite{shade1998layered, lsiTulsiani18, Shih3DP20} and multiplane image (MPI) \cite{szeliski1999stereo, zhou2018stereo, mildenhall2019llff, Srinivasan_2019_CVPR, Tucker_2020_CVPR}, become popular due to their appealing properties of explicitly modeling occluded contents. An MPI consists of multiple planes of RGB-$\alpha$ images at fixed depths, its performance is limited by sparse depths discretization. LDI stores multiple RGBD pixels at every pixel lattice, and thus naturally handles arbitrary number of layers. Rendering of LDI in \cite{lsiTulsiani18} leads to problems like cracks. More sophisticated LDI methods like \cite{Shih3DP20} contain iterative depth edges inpainting steps, which is prohibitively slow for real-world applications.

Another line of related works is self-supervised depth estimation. These works aim to train depth estimation models using image reconstruction error as the main supervision signal without ground truth depths. Exploiting epipolar geometry, \cite{Garg2016, deep3d2016, monodepth17} predict per-pixel disparities that recover one image from the other in the pair. \cite{monodepth17} additionally adds a left-right consistency term to improve the quality of the disparity maps. \cite{zhou2017unsupervised,sfmnet2017,mahjourian2018unsupervised,Wang_2018_CVPR,monodepth2} have been proposed to use monocular videos for self-supervision. Although depth estimation and view synthesis are closely related, good depth estimation results do not guarantee good view synthesis results, and vice versa. We show that our method achieves state-of-the-art performance in both view synthesis and depth estimation.

\vspace{-4mm}
\paragraph{Implicit 3D representations for view synthesis.} Recent works show that neural network can be used as an implicit representation for 3D shapes. To encode 3D shapes into the network weights, \cite{chen2018implicit_decoder, Jiang_2020_CVPR, Genova_2019_ICCV, atzmon2019controlling, occupancy2019, Park_2019_CVPR, Genova_2020_CVPR, Peng2020ECCV} map continuous 3D coordinates to signed distance functions or occupancy. However, they require supervision from ground truth 3D geometry. Others \cite{NEURIPS2019_b5dc4e5d, Niemeyer_2020_CVPR, yariv2020multiview} alleviate this requirements with differentiable rendering, which enables RGB-only supervision. Nonetheless, these methods do not deliver photo-realistic rendering results in scenes with complex structure. Recently, NeRF \cite{mildenhall2020nerf} shows astonishing results for novel view synthesis. NeRF works by mapping a continuous 3D coordinate and 2D viewing direction to a 4D output of RGB values and volume density. Works have been proposed to improve NeRF to images in the wild \cite{martinbrualla2020nerfw} and non-rigid scenes \cite{park2020deformable}. However, NeRF needs to be optimized per scene. PixelNerf \cite{yu2020pixelnerf} is proposed to solve the generalization problem while is does not solve the single-image scale ambiguity problem. GRF \cite{grf2020} is another improvement that works for multiple-view input. Neither of \cite{yu2020pixelnerf,grf2020} presents experiments on large scale real world datasets.

We take the best of the two worlds of NeRF and MPI and propose a new 3D representation, which we call MINE. Our method predicts planes of RGB-$\sigma$ images at any given arbitrary depths, thus allows continuous / dense 3D reconstruction of the scene. Unlike NeRF which encodes the scene geometry in the network weights, our network conditions on the input image, and thus can generalize to unseen scenes.

\section{Our Approach}  \label{sec_approach}
The input to our method is a single image, and the output is our 3D representation illustrated in Sec.~\ref{sec_appr_3drep}. Our network design and training pipeline are introduced in Sec.~\ref{sec_appr_network_training} and \ref{sec_appr_loss}. Furthermore, we discuss how our MINE relates to NeRF and MPI in Sec.~\ref{sec_appr_nerf} and \ref{sec_appr_mpi}. 

\subsection{3D Representation} \label{sec_appr_3drep}
\subsubsection{Planar Neural Radiance Field}
We utilize the perspective geometry to represent the camera frustum. Let us denote a pixel coordinate on the image plane as $[x, y]^\top\in \mathbb{R}^2$, and the pinhole camera intrinsic as $\text{K}\in \mathbb{R}^{3\times 3}$. A 3D point in the camera frustum is represented as $[x, y, z]^\top$, where $z$ is the depth of that point with respect to the camera. We define the conversion $\mathfrak{C}(\cdot)$ from perspective 3D coordinate $[x, y, z]^\top$ to Cartesian coordinate $[X, Y, Z]^\top$ as:
\begin{equation}
\small
    \mathfrak{C} (\begin{bmatrix} x\\ y\\ z\end{bmatrix})
    = \text{K}^{-1} \begin{bmatrix} x\\ y\\ 1\end{bmatrix} z 
    = \begin{bmatrix}
    f_x & 0 & c_x \\ 0 & f_y & c_y \\ 0 & 0 & 1
    \end{bmatrix}^{-1} \begin{bmatrix} zx\\ zy\\ z\end{bmatrix}.
\end{equation}

As shown in Fig.~\ref{fig_render}, we can sample arbitrary number of planes within the camera frustum with different depth values $z \in [z_n, z_f]$. Each plane consists of the RGB values $c_z: [x, y]^\top\to \mathbb{R}^3$ and volume densities $\sigma_z: [x, y]^\top \to \mathbb{R}^+$ of every point $[x, y]^\top \in \mathbb{R}^2$ on that plane. The volume density $\sigma(x,y,z)$ represents the differential probability of a ray terminating at an infinitesimal particle at location $[x,y,z]^\top$. The camera frustum within depth range $[z_n, z_f]$ is reconstructed continuously because the RGB $c(x, y, z)$ and $\sigma(x, y, z)$ of any position $[x,y,z]^\top$ are given by sampling a plane at depth $z$, and querying $c_{z}(x, y)$ and $\sigma_{z}(x, y)$. We call this the \textit{planar neural radiance field} because it represents the frustum using planes instead of rays in \cite{mildenhall2020nerf}.

\subsubsection{Volume Rendering} \label{sec_appr_render}
$c$ and $\sigma$ defined above are continuous two-dimensional functions that represent every possible position in the frustum. In practice, we discretize the planar radiance field in two aspects: a) The frustum consists of $N$ planes $\{c_{z_i}, \sigma_{z_i} \mid i=1,\cdots, N\}$. b) Each plane $(c_{z_i}, \sigma_{z_i})$ is simplified into a 4-channel image plane at depth $z_i$. Note the discretization is only for the convenience of rendering. The discretized representation is still able to acquire the RGB$\sigma$ values at any 3D position because: a) Each plane can be at arbitrary depth $z_i\in [z_n, z_f]$ and b) sub-pixel sampling is trivial at each 4-channel plane.

\vspace{-3mm}
\paragraph{Rendering the input image $\hat{\text{I}}_{src}$.} 
We first illustrate the rendering mechanism with the naive setting of rendering $\hat{\text{I}}_{src}$. A novel view can then be rendered in a similar way with an additional homography warping. Rendering $\hat{\text{I}}_{src}$ is straightforward using the principles from classical volume rendering \cite{kajiya1984ray,mildenhall2020nerf}, i.e.: \vspace{-2mm}
\begin{equation} \label{eq_render}
\small
\hat{\text{I}} = \sum_{i=1}^{N}T_i\big(1-\text{exp}(-\sigma_{z_i}\delta_{z_i})\big)c_{z_i}, \vspace{-2mm}
\end{equation}
\begin{equation*}
\text{where} \quad \quad \small
T_i = \text{exp}\bigg( -\sum_{j=1}^{i-1}\sigma_{z_j}\delta_{z_j} \bigg) : \mathbb{R}^2 \to \mathbb{R}^+ \quad \quad \quad \quad
\end{equation*}
is the map of accumulated transmittance from the first plane to plane $i$. Specifically, $T_i(x,y)$ denotes the probability of a ray travels from $(x,y,z_1)$ to $(x,y,z_i)$ without hitting any object. Furthermore, 
\begin{equation*}
\small
\delta_{z_i}(x, y) = \|\mathfrak{C}([x,y,z_{i+1}]^\top) - \mathfrak{C}([x,y,z_{i}]^\top)\|_2 : \mathbb{R}^2 \to \mathbb{R}^+ 
\end{equation*}
is the distance map between plane $i+1$ and $i$. 

According to Eq.~\ref{eq_render}, the collection of $\{(c_{z_i}, \sigma_{z_i}, z_i) \mid i=1,\cdots,N\}$ is required to render the input image. As shown in Sec.~\ref{sec_appr_network_training}, $(c_{z_i}, \sigma_{z_i})$ is the output of our network, which takes $\text{I}_{src}$ and $d_i=1 / z_i$ as inputs. Following the stratified sampling strategy of \cite{mildenhall2020nerf}, $\{z_i \mid i=1,\cdots,N\}$ is sampled within $[z_n, z_f]$. In fact, we sample disparity $\{d_i=1/z_i\}$ in the perspective geometry. Specifically, $[d_n, d_f]$ is partitioned into $N$ evenly spaced bins, and a sample is drawn uniformly from each bin, i.e.: 
\begin{equation} \label{eq_disparity_sample}
\small
    d_i \sim \mathcal{U}\bigg[ d_n + \frac{i}{N}(d_f - d_n), d_n + \frac{i-1}{N}(d_f - d_n)\bigg].
\end{equation}
The sampling strategy ensures that our network is exposed to every depth value in the frustum during training, and thus learning a continuous $(c_{z_i}, \sigma_{z_i})$.

In addition, the depth map of the input image can be rendered in a similar way as Eq.~\ref{eq_render}, i.e.: \vspace{-2mm}
\begin{equation} \label{eq_render_depth}
    \hat{\mathcal{Z}} = \sum_{i=1}^{N}T_i\big(1-\text{exp}(-\sigma_{z_i}\delta_{z_i})\big) z_i.
    \vspace{-3mm}
\end{equation} 
\begin{figure*}[t!] \centering
\includegraphics[width=0.99\textwidth]{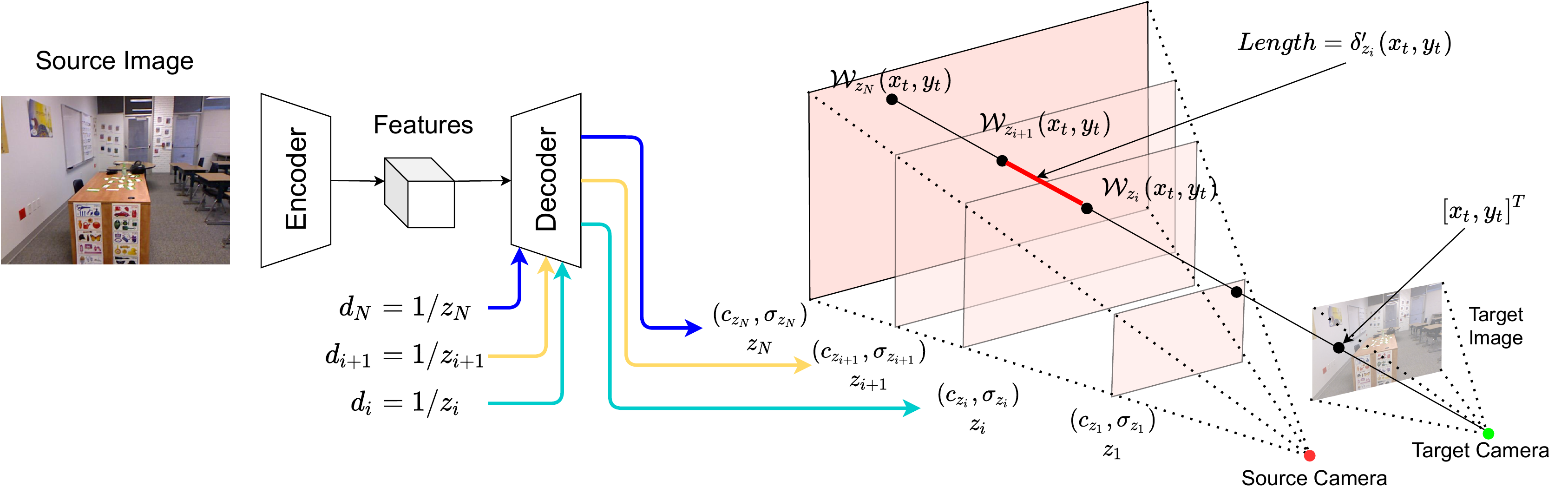}
\caption{Our network is an encoder-decoder architecture (c.f.  Sec.~\ref{sec_appr_network_training}) that takes an input image and outputs the reconstructed source camera frustum. We then render the reconstructed source camera frustum into a novel view (c.f. Sec.~\ref{sec_appr_render}).}
\label{fig_render}
\vspace{-8pt}
\end{figure*}
\vspace{-5mm}
\paragraph{Rendering a novel view $\hat{\text{I}}_{tgt}$.}
As illustrated in Fig.~\ref{fig_render}, the rendering into a novel view with camera rotation $\text{R}\in \mathbb{R}^{3\times3}$ and translation $\textbf{t}\in \mathbb{R}^3$ is achieved in three steps: 
\vspace{-4mm}
\paragraph{1)} Apply a homography warping $\mathcal{W}(.)$ to establish the correspondence between the source pixel coordinates $[x_s, y_s]^\top$ and the target pixel coordinates $[x_t, y_t]^\top$. 
We follow the standard inverse homography \cite{Hartley2004, zhou2018stereo, Tucker_2020_CVPR} to define $\mathcal{W}(.)$. The correspondence between a pixel coordinate $[x_t, y_t]^\top$ in the target plane and and a pixel coordinate $[x_s, y_s]^\top$ in the source plane is given by:
\begin{equation} \label{eq_homography}
\small
    \begin{bmatrix} x_s, y_s, 1\end{bmatrix}^\top \sim \text{K}\bigg( \text{R} - \frac{\textbf{t} \mathbf{n^\top}}{z_i} \bigg)\text{K}^{-1}
    \begin{bmatrix} x_t, y_t, 1\end{bmatrix}^\top,
    \vspace{-2mm}
\end{equation}
where $\mathbf{n}=[0, 0, 1]^\top$ is the normal vector of the fronto-parallel plane $(c_{z_i}, \sigma_{z_i})$ with respect to the source camera. For brevity, we now denote the above warping as $[x_s, y_s]^\top=\mathcal{W}_{z_i}(x_t, y_t)$ for the plane at depth $z_i$ with respective to the source camera.
We then compute the plane projections $(c'_{z_i}, \sigma'_{z_i})$ at the target frame as:
    $c'_{z_i}(x_t, y_t) = c_{z_i}(x_s, y_s),$ and $\sigma'_{z_i}(x_t, y_t) = \sigma_{z_i}(x_s, y_s)$.
Note that the $N$ planes are fronto-parallel to the source camera, and therefore $(c'_{z_i}, \sigma'_{z_i})$ is just the projection into the target camera.

\vspace{-4mm}
\paragraph{2)} 
The volume rendering relies on the density $\sigma$ at each location, and the distances between each point along the ray. Consequently, we can compute:
\begin{equation}\label{eq_delta_prime}
\begin{split}
    \delta'_{z_i}(x_t, y_t) = \| & \mathfrak{C}([\mathcal{W}_{z_{i+1}}(x_t, y_t),z_{i+1}]^\top) \\
    & - \mathfrak{C}([\mathcal{W}_{z_i}(x_t, y_t),z_{i}]^\top)\|_2.
\end{split}
\end{equation}
As illustrated in Fig.~\ref{fig_render}, let us imagine a ray that starts from target camera origin and intersects the target image at pixel coordinate $[x_t, y_t]^\top$ to better understand Eq.~\ref{eq_delta_prime}. This ray intersects the $(c_{z_i}, \sigma_{z_i})$ plane at pixel coordinate $\mathcal{W}_{z_i}(x_t, y_t)$ with respect to the source camera. Similarly, the ray intersects the $(c_{z_{i+1}}, \sigma_{z_{i+1}})$ plane at source camera pixel coordinate $\mathcal{W}_{z_{i+1}}(x_t, y_t)$. $\delta'_{z_i}(x_t, y_t)$ represents the Euclidean distance between the two intersections.

\vspace{-4mm}
\paragraph{3)} Finally, the rendering into a novel view can be achieved by applying Eq.~\ref{eq_render} after replacing $c,\sigma,\delta$ with $c',\sigma',\delta'$.

\subsection{Network and Training Design} \label{sec_appr_network_training}
As shown above, a discretized planar radiance field requires a set of depth samples $\{z_i \mid i=1,\cdots,N\}$, and 4-channel images $\{(c_{z_i}, \sigma_{z_i}) \mid i=1,\cdots,N\}$. Depth samples $\{z_i\}$ or disparity samples $\{d_i=1/z_i\}$ are randomly sampled according to Eq.~\ref{eq_disparity_sample}.

\vspace{-4mm}
\paragraph{Encoder-Decoder Structure.}
The 4-channel images $\{(c_{z_i}, \sigma_{z_i})\}$ are predictions from our network, which takes a single image and $\{z_i\}$ as input. Our network architecture is shown in Fig.~\ref{fig_render}. The encoder takes the image as input and produces a series of feature maps. We utilize the Resnet-50 \cite{resnet2016} as the encoder. The decoder takes the feature maps and a single disparity value $d_i=1/z_i$ as input, and produces the 4-channel image $(c_{z_i}, \sigma_{z_i})$. The decoder design is similar to Monodepth2 \cite{monodepth2}. In training and inference, the encoder runs only once per image (or per mini-batch of images), while the decoder runs $N$-times to generate the discrete set of planes $\{(c_{z_i}, \sigma_{z_i}) \mid i=1,\cdots,N\}$.

\vspace{-4mm}
\paragraph{Disparity Encoding.} We find that directly feeding $d_i$ into the decoder gives poor performance, which is consistent with \cite{mildenhall2020nerf, rahaman2019spectral, NEURIPS2020_fourier_feature}. To circumvent this problem, we apply an encoding function $\gamma: \mathbb{R}\to \mathbb{R}^L$ to $d_i$ before feeding into the decoder, i.e.:
\begin{equation}
\begin{split}
    \gamma(d_i) = [ & \sin{(2^0\pi d_i)}, \cos{(2^0\pi d_i)}, \cdots, \\ 
    & \sin{(2^{L-1}\pi d_i)}, \cos{(2^{L-1}\pi d_i)}].
\end{split}
\end{equation}

\subsection{Supervision with RGB Videos} \label{sec_appr_loss}
Multi-view images or RGB videos are used to train the network similar to \cite{Tucker_2020_CVPR}. During training, an input image $\text{I}_{src}$ is fed into the network and then rendered into $(\hat{\text{I}}_{tgt},~\hat{\mathcal{Z}}_{tgt})$, according to the novel view camera rotation $\text{R}$ and scale-calibrated camera translation $\textbf{t}'$. The core supervision is by comparing $\hat{\text{I}}_{tgt}$ with the ground truth target image $\text{I}_{tgt}$.

\subsubsection{Scale Calibration}\label{sec_scale_calib}
The depth scale is ambiguous up to a scale factor $s\in \mathbb{R}^+$ since the input to our system is a single image. The range of the frustum reconstruction $[z_n, z_f]$ is pre-defined as a hyper-parameter, which we set as $z_n=1, \,z_f=1000$. Instead of scaling our 3D representation, we scale the camera translation $\textbf{t}$ into $\textbf{t}'$ at both training and inference. 

To solve the scale factor $s$, we perform a scale calibration between the sparse 3D points from video Structure-from-Motion (SfM) and our synthesized depth map of Eq.~\ref{eq_render_depth}. Specifically, we run the SfM using COLMAP \cite{schoenberger2016sfm, schoenberger2016mvs} on each video to get a sparse point set $\mathbf{P_s}=\{(x_j,y_j,z_j)\}$ for each image. The coordinates here follow the same perspective geometry, i.e. $[x_j, y_j]^\top$ is the pixel coordinate on the image, and $z_j$ is the depth of the corresponding 3D point. After feeding the source image to our network and rendering the predicted depth map $\hat{\mathcal{Z}}_{src}$ using Eq.~\ref{eq_render_depth}, similar to \cite{Tucker_2020_CVPR}, the scale is estimated by:
\begin{equation}
s = \exp{\bigg[ \frac{1}{|\mathbf{P_s}|} \sum_{(x,y,z)\in \mathbf{P_s}} \big(\ln(\hat{\mathcal{Z}}_{src}(x,y) - \ln{z}\big) \bigg]}.
\end{equation}
Finally, the calibrated translation is given by $\textbf{t}' = \textbf{t} \cdot s$.

\vspace{-3mm}
\subsubsection{Loss Functions}
There are four terms in the loss function: RGB L1 loss $\mathcal{L}_{\text{L1}}$, RGB SSIM loss $\mathcal{L}_{\text{ssim}}$, edge-aware disparity map smoothness loss $\mathcal{L}_{\text{smooth}}$, and the optional sparse disparity loss $\mathcal{L}_{d}$. The total loss is given by:
\begin{equation}
\small
    \mathcal{L} = \lambda_{\text{L1}}\mathcal{L}_{\text{L1}} + \lambda_{\text{ssim}}\mathcal{L}_{\text{ssim}} + \lambda_{\text{smooth}}\mathcal{L}_{\text{smooth}} + \lambda_{d}\mathcal{L}_{d},
\end{equation}
where $\lambda_{\text{L1}}$, $\lambda_{\text{ssim}}$, $\lambda_{\text{smooth}}$ and $\lambda_{d}$ are hyperparameters to weigh the respective loss term.

\vspace{-3mm}
\paragraph{RGB L1 and SSIM loss.}
The L1 and SSIM \cite{wang2004image} losses:
\begin{equation}
\small
    \mathcal{L}_{\text{L1}} = \frac{1}{3HW}\sum |\hat{\text{I}}_{tgt} - \text{I}_{tgt}|, \quad
    \mathcal{L}_{\text{ssim}} = 1 - \text{SSIM}(\hat{\text{I}}_{tgt}, \text{I}_{tgt})
\end{equation}
are to encourage the synthesized target image $\hat{\text{I}}_{tgt}$ to match the ground truth $\text{I}_{tgt}$. Both $\hat{\text{I}}_{tgt}$ and $\text{I}_{tgt}$ are 3-channel RGB images of size $H\times W$.

\vspace{-3mm}
\paragraph{Edge-aware disparity map smoothness loss.}
We impose an edge-aware smoothness loss on the synthesized disparity map to penalize drastic changes in disparities at locations where the original image is smooth, and to align edges in the disparity maps and original images correctly. Note that there are many forms of such a loss \cite{monodepth17, monodepth2, Wang_2018_CVPR, Tucker_2020_CVPR}, we adopt the one in \cite{monodepth17, monodepth2}, which is defined as: 
\begin{equation}
    \mathcal{L}_{smooth} = |\partial{_{x}}\hat{\mathcal{D}}^{*}|\exp^{-|\partial{_{x}}\text{I}|} + |\partial{_{y}}\hat{\mathcal{D}}^{*}|\exp^{-|\partial{_{y}}\text{I}|}, 
\end{equation}
where $\partial{_{x}}$ and $\partial{_{y}}$ are the image gradients, and  $\hat{\mathcal{D}}^{*} = \hat{\mathcal{D}} / \bar{\mathcal{D}}$ is the mean-normalized disparity, where $\hat{\mathcal{D}}=1/\hat{\mathcal{Z}}$.

\vspace{-3mm}
\paragraph{Sparse disparity loss.}
In the case that SfM is adopted to pre-process the input images/videos to solve the scale ambiguity, we apply the sparse disparity loss to facilitate the depth/disparity predictions. Nonetheless, note that this is optional. In particular, SfM is not necessary and the sparse disparity loss is not applicable in datasets such as KITTI, where the scale is fixed to $s=1$. We follow the log disparity style as \cite{Tucker_2020_CVPR,eigen2014depth}. 
\begin{equation}
\small
\begin{split}
    & \mathcal{L}_d = 0.5\mathcal{L}_d^{src} + 0.5 \mathcal{L}_d^{tgt}, \quad \text{where} \\
    & \mathcal{L}_d^{src} = \frac{1}{|\mathbf{P_s}|}\sum_{(x,y,z)\in\mathbf{P_s}}\bigg( \ln{\frac{\hat{\mathcal{D}}_{src}(x,y)}{s}} - \ln{\frac{1}{z}} \bigg),\\
    & \mathcal{L}_d^{tgt} = \frac{1}{|\mathbf{P_t}|}\sum_{(x,y,z)\in\mathbf{P_t}}\bigg( \ln{\frac{\hat{\mathcal{D}}_{tgt}(x,y)}{s}} - \ln{\frac{1}{z}} \bigg).
\end{split}
\end{equation}
Note that we need to scale the disparity maps because the translation $\textbf{t}$ is calibrated with $s$ as shown in Sec.~\ref{sec_scale_calib}. The translation and depth should be calibrated together.

\subsection{Our Relation to NeRF} \label{sec_appr_nerf}
Our MINE shares similar underlying representation, i.e., RGB and volume density at arbitrary position in the space. \textbf{Advantages}: 1) Our MINE generalizes to unseen scenes, while NeRF has to be optimized per scene. 2) To render a novel view, our MINE requires lesser network inferences (e.g. 32 network inferences), while NeRF requires millions of network inferences. \textbf{Limitations}: 1) Our MINE takes only one image as input and therefore it is impossible to reconstruct the whole object from all 360\degree. 2) MINE does not take viewing direction as input, therefore, it is not able to model complex view-dependent effects. 



\subsection{Our Relation to MPI} \label{sec_appr_mpi}

The MPI representation in \cite{Tucker_2020_CVPR} is a special case of our representation described in Sec.~\ref{sec_appr_3drep}. 

\begin{proof}
Instead of sampling with Eq.~\ref{eq_disparity_sample}, we can simply set $d_i$ as the bin edges given by:
\begin{equation} \label{eq_fixed_disparity}
    d_i = d_n + (i-1)/N\cdot (d_f - d_n).
   \vspace{-1mm}
\end{equation}
Additionally, we can define the alpha map at depth $z_i$ as $\alpha_{z_i}: \mathbb{R}^2 \to \mathbb{R}^+$ shown in Eq.~\ref{eq_alpha_sigma_relation}. Now, Eq.~\ref{eq_render} can be rewritten into:
\begin{equation} \label{eq_render_mpi}
\small
    \hat{\text{I}} = \sum_{i=1}^{N}T_i\alpha_{z_i}c_{z_i},~\text{\normalsize{where}} \quad T_i=\prod_{j=1}^{i-1}(1-\alpha_{z_j}),
    \vspace{-2mm}
\end{equation} 
which is identical to the MPI compositing operation in \cite{Tucker_2020_CVPR} (c.f. Eq.~5 in \cite{Tucker_2020_CVPR}), where they directly predict: 
\begin{equation} \label{eq_alpha_sigma_relation}
    \alpha_{z_i} = 1 - \text{exp}(-\sigma_{z_i}\delta_{z_i})
    \vspace{-1mm}
\end{equation}
instead of $\sigma_{z_i}$. Note that  $\delta_{z_i}$ is now a constant since $d_i$ is set as the bin edges without random sampling.
\end{proof}
%
%
%
\subsection{Our Relation to pixelNeRF and GRF} \label{sec_appr_pixelnerf_grf}
Our MINE is different from pixelNeRF \cite{yu2020pixelnerf} and GRF \cite{grf2020} by: \textbf{(a)} MINE directly models the frustum of the source camera, while both pixelNeRF and GRF model the entire 3D space. \textbf{(b)} MINE reconstructs the frustum of the source camera per plane, while pixelNeRF and GRF reconstruct the entire 3D space per ray. \textbf{(c)} Neither pixelNeRF nor GRF presents experiments on large scale real world datasets, while MINE presents results on large scale indoor / outdoor datasets, namely RealEstate10K, NYUv2 and KITTI.

A direct consequent of (a) (b) is that our MINE is significantly more efficient at inference. 
Both pixelNeRF and GRF render the output image pixel by pixel, and therefore the number of forward passes required is proportional to the spatial resolution of the output, the number of points along each ray, and the number of target views to render. 
On the contrary, since our MINE reconstructs the entire frustum of the source camera per plane, we only require $N_\text{{planes}}$ forward passes of the fully-convolutional decoder to obtain the representation. The rendering for any novel view only requires an additional homography warping step. 
Detailed analysis is presented in the supplementary materials.

\section{Experiments}  \label{sec_experiments}
\begin{table*}[]
\small
\centering
{
\begin{tabular}{c|cccc | ccc}
\hline
                 & Train Res. & $N$ & Pre-trained & Depth Smoothess & LPIPS$\downarrow$ & SSIM$\uparrow$ & PSNR$\uparrow$ \\
\hline 
MINE             & 384x128    & 32  & N            & Y                & 0.129             & 0.812          & 21.4           \\
MINE             & 384x128    & 32  & Y            & N                & 0.123             & 0.816          & 21.6           \\
MINE             & 384x128    & 32  & Y            & Y                & 0.122             & 0.815          & 21.6           \\
MINE             & 384x128    & 64  & Y            & Y                & 0.117             & 0.818          & 21.6           \\
MINE             & 384x128    & 256 & Y            & Y                & \textbf{0.112}    & \textbf{0.828} & \textbf{21.9}  \\
\hline
Tulsiani et. al. \cite{lsiTulsiani18} & 768x256    & NA  & NA           & NA               & -                 & 0.572          & 16.5           \\
MPI \cite{Tucker_2020_CVPR}              & 768x256    & 32  & NA           & NA               & -                 & 0.733          & 19.5           \\
MINE             & 768x256    & 32  & Y            & Y                & 0.112             & \textbf{0.822} & 21.4           \\
MINE             & 768x256    & 64  & Y            & Y                & \textbf{0.108}    & 0.820          & 21.3 \\
\hline
\end{tabular}
}
\vspace{1pt}
\caption{View synthesis on KITTI dataset. Note that \cite{lsiTulsiani18} trains the model at $768\times256$ and tests at $384\times128$ to avoid cracks in the output, and \cite{Tucker_2020_CVPR} adopts this setting for comparison. We follow this setting and all our models are tested with resolution of $384\times128$.}\label{table_kitti_results}
\end{table*}
\begin{figure*}[t!] \centering
\setlength{\tabcolsep}{0pt}
\renewcommand{\arraystretch}{0.0}
\begin{tabular}{lccc}
     \raisebox{2\totalheight}{Input} &  {\includegraphics[width=0.29\textwidth]{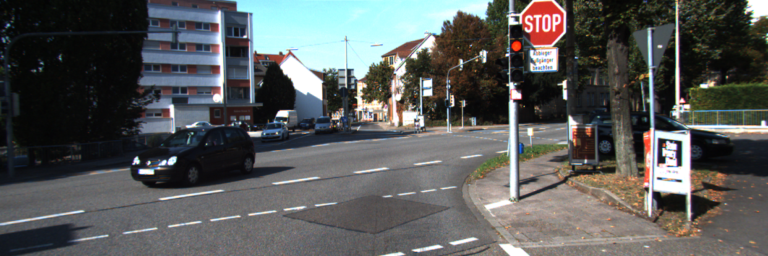}} &  {\includegraphics[width=0.29\textwidth]{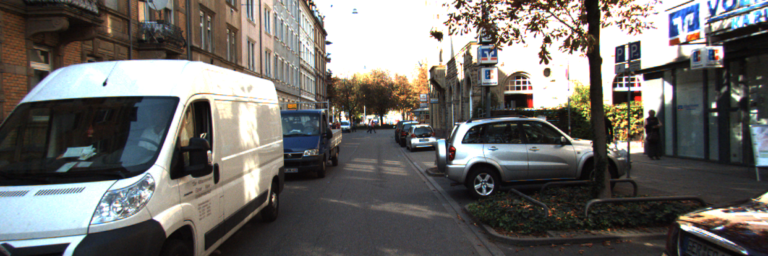}} &  {\includegraphics[width=0.29\textwidth]{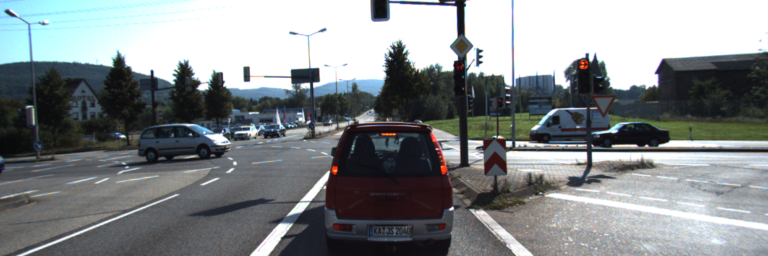}} \tabularnewline
     
     \raisebox{2\totalheight}{Target GT\,\,\,} &  {\includegraphics[width=0.29\textwidth]{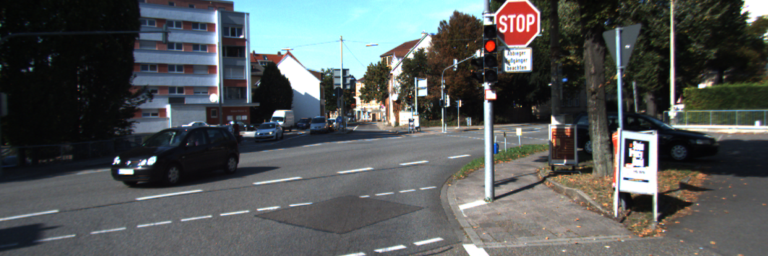}} &  {\includegraphics[width=0.29\textwidth]{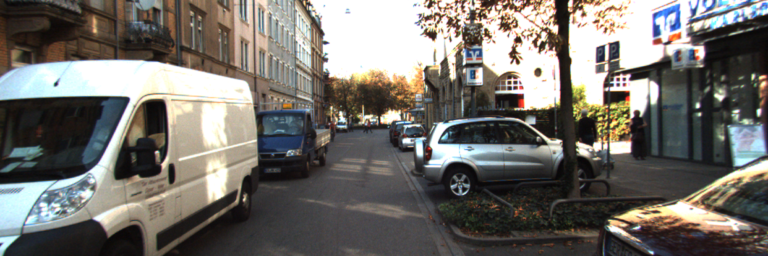}} &  {\includegraphics[width=0.29\textwidth]{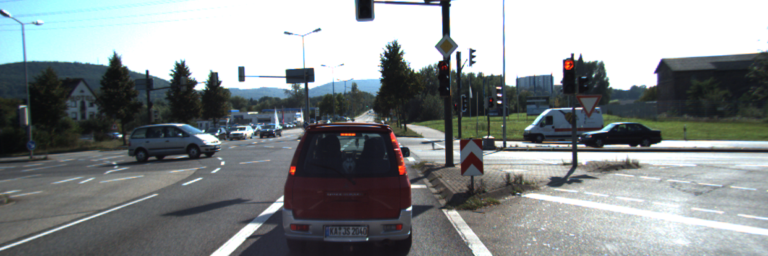}} \tabularnewline
     
     \raisebox{2\totalheight}{MINE} &  {\includegraphics[width=0.29\textwidth]{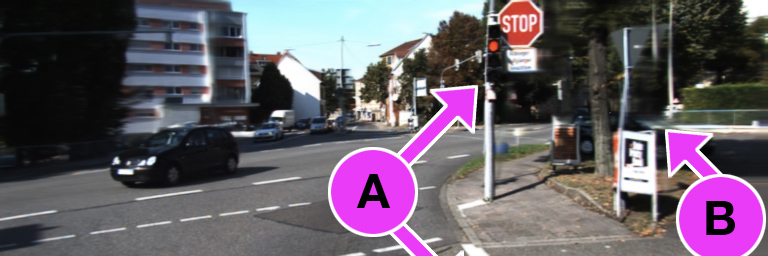}} &  {\includegraphics[width=0.29\textwidth]{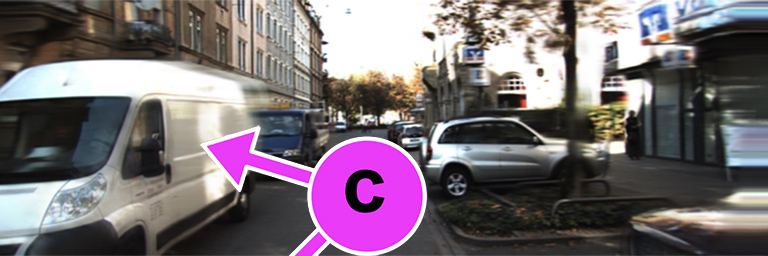}} &  {\includegraphics[width=0.29\textwidth]{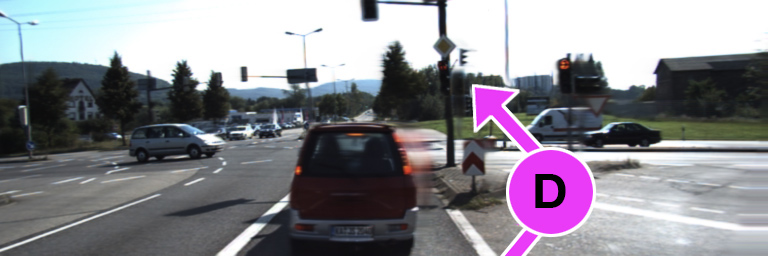}} \tabularnewline
     
     \raisebox{2\totalheight}{MPI \cite{Tucker_2020_CVPR}} &  {\includegraphics[width=0.29\textwidth]{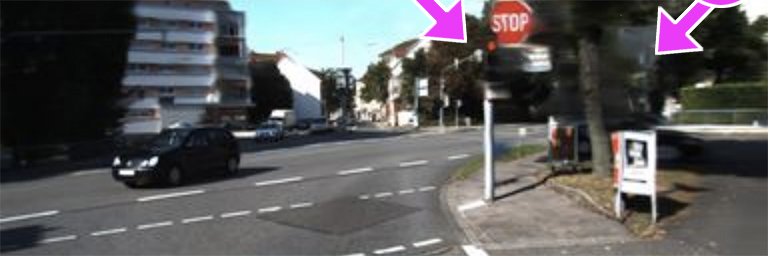}} &  {\includegraphics[width=0.29\textwidth]{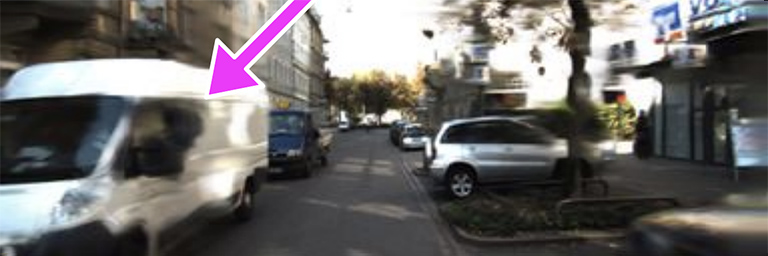}} &  {\includegraphics[width=0.29\textwidth]{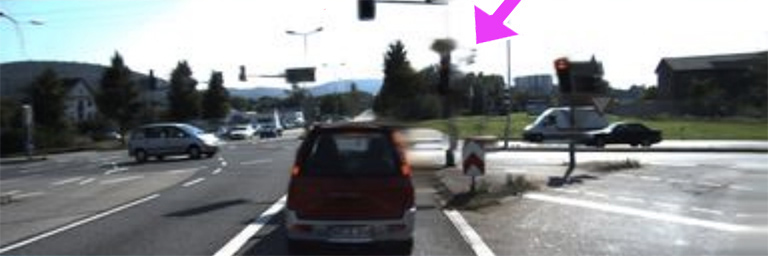}}
\end{tabular}
    \vspace{1pt}
    \caption{Qualitative comparison on KITTI. Note that these examples are not cherry-picked, they are the same images used in \cite{Tucker_2020_CVPR}.} 
    \label{fig_kitti_vis}
    \vspace{-4pt}
\end{figure*} 
For novel view synthesis, we perform both quantitative and qualitative comparisons with state-of-the-art methods on the RealEstate10K~\cite{zhou2018stereo}, flowers light field \cite{Srinivasan_2017_ICCV}, and KITTI \cite{Geiger2013IJRR} datasets. To measure the quality of the generated images, we compute the Structural Similarity Index (SSIM) \cite{1284395}, PSNR, and the recently proposed LPIPS perceptual similarity \cite{zhang2018perceptual}. We use an ImageNet-trained VGG16 \cite{Simonyan15} model when computing the LPIPS score.
For depth estimation from single image, we perform evaluations on the iBims-1 {\cite{Koch18:ECS}} dataset and the NYU-Depth V2 {\cite{Silberman:ECCV12}} dataset.

\subsection{View Synthesis on KITTI}
%
Following the settings of \cite{lsiTulsiani18,Tucker_2020_CVPR}, we train our models on the 20 city sequences from the KITTI Raw dataset \cite{Geiger2013IJRR}, and evaluates on another 4 city sequences. We fix the scale factor to 1 because only stereo pairs with a constant scale are utilized. During training, the left or right image is randomly taken as the source image and the target image. Following \cite{Tucker_2020_CVPR}, we crop 5\% from all sides of all images before computing the scores in testing.
Quantitative comparisons with \cite{Tucker_2020_CVPR,lsiTulsiani18} are presented in Tab. \ref{table_kitti_results}. Both our 32- and 64-plane models outperform these existing methods by a large margin. Notably, we significantly improve the SSIM from 0.733 to 0.822 when compared to \cite{Tucker_2020_CVPR}.
We also qualitatively demonstrate our superior view synthesis performance in Fig.~\ref{fig_kitti_vis}. Compared to \cite{Tucker_2020_CVPR}, we generate more realistic images with lesser artefacts and shape distortions. The visualization verifies our ability to model the geometry and texture of complex scenes.

\begin{figure}[h!] \centering
\setlength{\tabcolsep}{0pt}
\renewcommand{\arraystretch}{0.0}
\resizebox{\columnwidth}{!}{
\begin{tabular}{c@{\hspace{0.1cm}}cc}
     RGB & \includegraphics[width=0.26\textwidth]{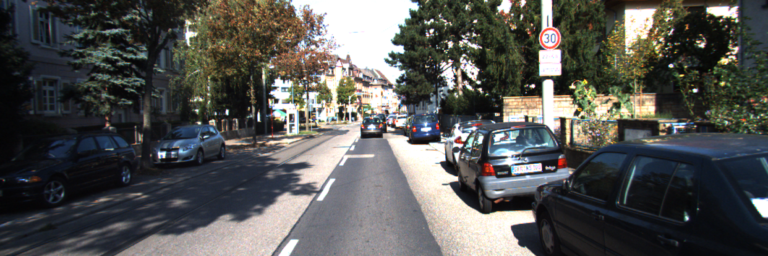} & \includegraphics[width=0.26\textwidth]{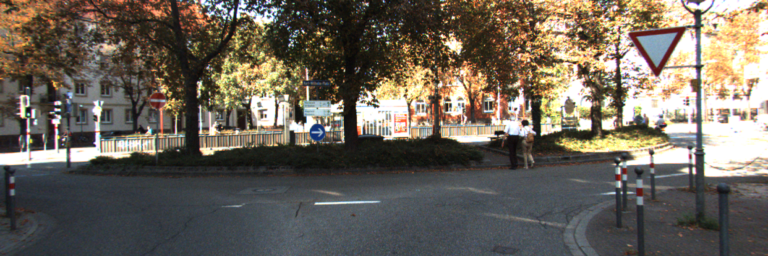} \\
     nosmooth & \includegraphics[width=0.26\textwidth]{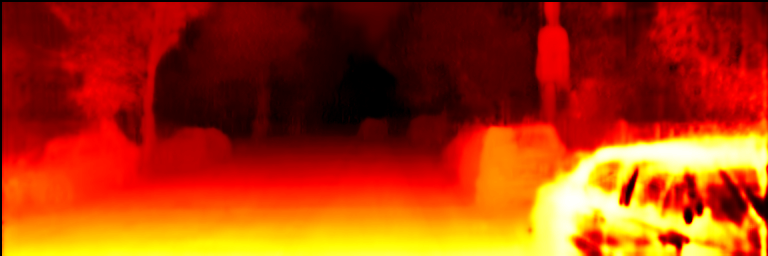} & \includegraphics[width=0.26\textwidth]{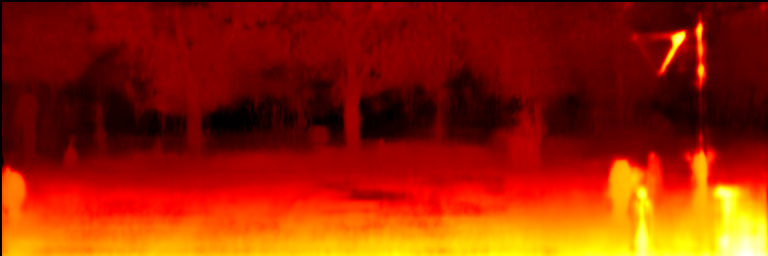} \\
     full & \includegraphics[width=0.26\textwidth]{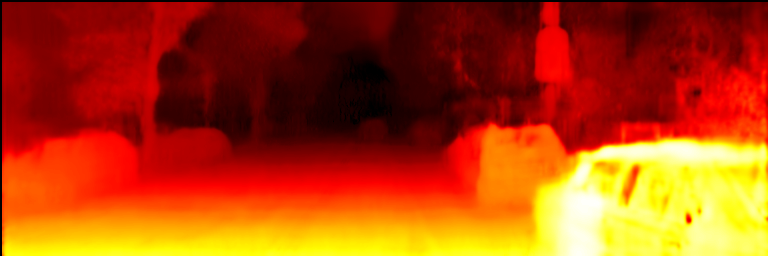} & \includegraphics[width=0.26\textwidth]{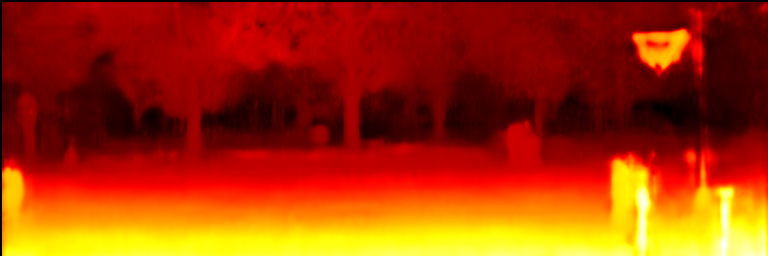} \\
\end{tabular}}
\vspace{1pt}
\caption{Effects of the edge-aware smoothness loss on KITTI.} 
\vspace{-8pt}
\label{fig_kitti_edgeaware_vis}
\end{figure} 
\vspace{-4mm}
\paragraph{Ablation Studies.} As shown in Tab.~\ref{table_kitti_results}, ablation studies are conducted on the KITTI dataset to validate some design choices. We observe that ImageNet pre-training for the encoder brings moderate improvements on all metrics. The edge-aware depth smoothness loss only brings marginal improvements quantitatively, but we qualitatively show in Fig.~\ref{fig_kitti_edgeaware_vis} that it enables the model to synthesize better disparity maps. More importantly, we see consistent improvements with increasing $N$. Since the model capacity remains the same as we vary $N$, the improvements can be attributed to the ability of the model to learn more complex scene geometry when sampling the depths more densely 
while training.

\subsection{View Synthesis on RealEstate10K}
%
%
%
\begin{table*}[t!]
\centering
\resizebox{\textwidth}{!}
{
\small
\begin{tabular}{c | ccc | ccc| ccc}
\hline
              & \multicolumn{3}{c|}{LPIPS$\downarrow$}   & \multicolumn{3}{c|}{SSIM$\uparrow$}     & \multicolumn{3}{c}{PSNR$\uparrow$}    \\
Method        & $n = 5$  & $n = 10$ & $n = random$ & $n = 5$  & $n = 10$ & $n = random$ & $n = 5$ & $n = 10$ & $n = random$ \\
\hline
SynSin \cite{Wiles_2020_CVPR}       & - & - & -     & - & - & 0.74     & - & - & 22.31      \\
MPI \cite{Tucker_2020_CVPR}           & 0.0967 & 0.1420 & 0.1761     & 0.8699 & 0.8124 & 0.7851     & 27.05 & 24.43  & 23.52      \\
MINE ($N = 32$) & 0.0934 & 0.1346 & 0.1674     & 0.8970 & 0.8464 & 0.8172     & \textbf{28.51} & \textbf{25.73}  & \textbf{24.56}      \\
MINE ($N = 64$) & \textbf{0.0896} & \textbf{0.1280} & \textbf{0.1562} & \textbf{0.8974} & \textbf{0.8500} & \textbf{0.8219} & 28.39 & 25.71  & 24.50      \\
\hline
\end{tabular}
}
\caption{Results on RealEstate10K \cite{zhou2018stereo}. $\uparrow$ denotes higher is better and $\downarrow$ means otherwise. $n$ is the number of frames between the source and target frames. 
The results of SynSin are from the original paper, where they use the same test setting (target frames are chosen randomly from within 30 frames of the source frames) as ours, but a different set of test pairs. They also use a lower resolution of 256x256.}\label{table:realestate_results}
\vspace{-2pt}
\end{table*}
\begin{table*}[t]
\setlength{\tabcolsep}{3pt}
\resizebox{\textwidth}{!}{
\begin{tabular}{ccc | cccccc | cccccc}
\hline
    &    &   & \multicolumn{6}{c|}{NYU-Depth V2 \cite{Silberman:ECCV12}}  & \multicolumn{6}{c}{iBims-1 \cite{Koch18:ECS}}  \\ 
Method        & Supervision & Dataset            & rel$\downarrow$           & log10$\downarrow$         &
RMS$\downarrow$          & $\sigma$1$\uparrow$        & $\sigma$2$\uparrow$       & $\sigma$3$\uparrow$     & rel$\downarrow$          & log10$\downarrow$         & RMS$\downarrow$           & $\sigma$1$\uparrow$       & $\sigma$2$\uparrow$        & $\sigma$3$\uparrow$        \\
\hline
DIW \cite{NIPS2016_0deb1c54}           & Depth       & DIW                & 0.25          & 0.1           & 0.76         & 0.62          & 0.88          & 0.96       & 0.25         & 0.1           & 1             & 0.61         & 0.86          & 0.95          \\
DIW \cite{NIPS2016_0deb1c54}           & Depth       & DIW+NYU            & 0.19          & 0.08          & 0.6          & 0.73          & 0.93          & 0.98       & 0.19         & 0.08          & 0.8           & 0.72         & 0.91          & 0.97          \\
MegaDepth \cite{MegaDepthLi18}     & Depth       & Mega               & 0.24          & 0.09          & 0.72         & 0.63          & 0.88          & 0.96       & 0.23         & 0.09          & 0.83          & 0.67         & 0.89          & 0.96          \\
MegaDepth \cite{MegaDepthLi18}     & Depth       & Mega+DIW           & 0.21          & 0.08          & 0.65         & 0.68          & 0.91          & 0.97       & 0.2          & 0.08          & 0.78          & 0.7          & 0.91          & 0.97          \\
3DKenBurns \cite{Niklaus_TOG_2019}    & Depth       & Mega+NYU+3DKenBurn & \textbf{0.08} & \textbf{0.03} & \textbf{0.3} & \textbf{0.94} & \textbf{0.99} & \textbf{1} & \textbf{0.1} & \textbf{0.04} & \textbf{0.47} & \textbf{0.9} & \textbf{0.97} & \textbf{0.99} \\
MiDaS v2.1 \cite{Ranftl2020} & Depth & MiDaS 10 datasets & 0.16 & 0.06 & 0.50 & 0.80 & 0.95 & 0.99 & 0.14 & 0.06 & 0.57 & 0.84 & \textbf{0.97} & \textbf{0.99} \\
\hline
MPI \cite{Tucker_2020_CVPR}           & RGB   & RealEstate10K      & 0.15          & 0.06          & 0.49         & 0.81          & 0.96          & 0.99       & 0.21         & 0.08          & 0.85          & 0.7          & 0.91          & 0.97          \\
MINE ($N = 64$) & RGB   & RealEstate10K      & 0.11          & 0.05          & 0.40         & 0.88          & 0.98          & 0.99       & 0.11         & 0.05          & 0.53          & 0.87         & \textbf{0.97} & \textbf{0.99} \\
\hline
\end{tabular}}
\caption{Depth estimation results on iBims-1 and NYU-Depth V2. We significantly outperform MPI \cite{Tucker_2020_CVPR} which is also supervised on only RGB images and sparse depth, and achieves comparable performance with state-of-the-art methods that use dense depth supervision.}
\label{table_depth_estimation}
\vspace{-2pt}
\end{table*}
\begin{figure*}[t!] \small \centering
\begin{tabular}{ccccc}
     Input & Target GT & MPI \cite{Tucker_2020_CVPR} & MINE ($N = 64$) \\ 
     \includegraphics[width=0.20\textwidth]{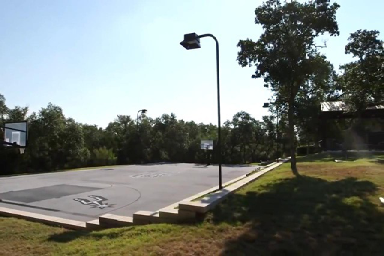} & \includegraphics[width=0.20\textwidth]{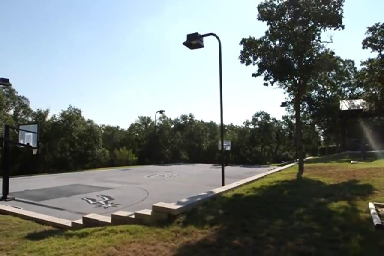} & \includegraphics[width=0.20\textwidth]{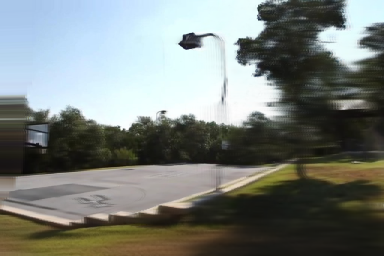} & \includegraphics[width=0.20\textwidth]{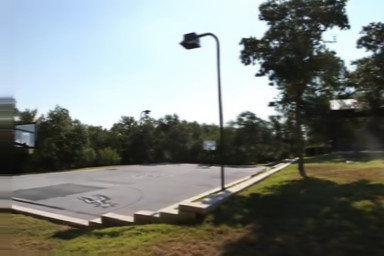} \\
     \includegraphics[width=0.20\textwidth]{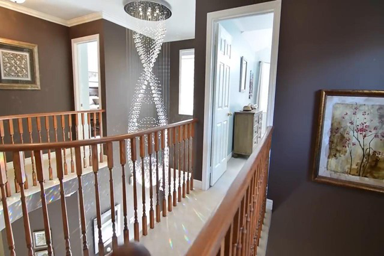} & \includegraphics[width=0.20\textwidth]{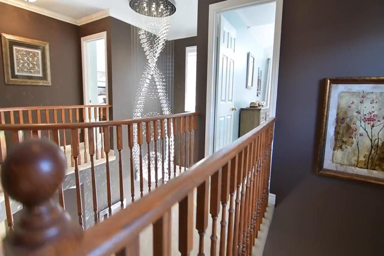} & \includegraphics[width=0.20\textwidth]{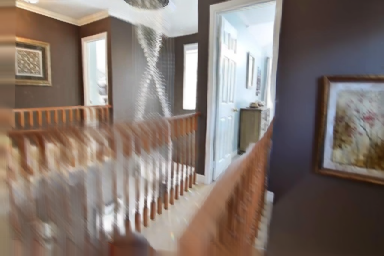} & \includegraphics[width=0.20\textwidth]{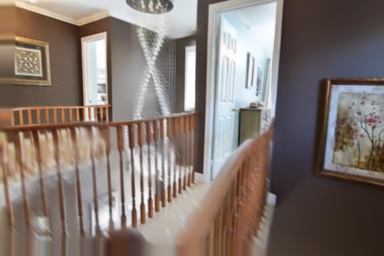} \\
     \includegraphics[width=0.20\textwidth]{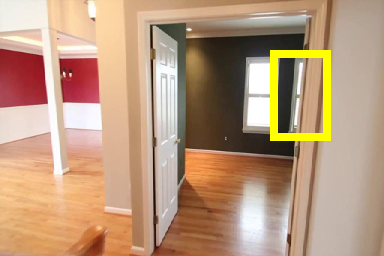} & \includegraphics[width=0.20\textwidth]{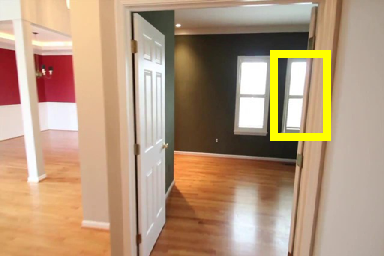} & \includegraphics[width=0.20\textwidth]{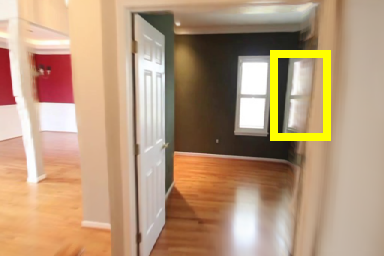} & \includegraphics[width=0.20\textwidth]{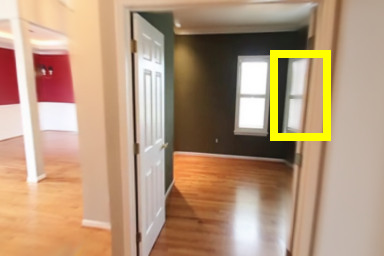}
\end{tabular}
    \caption{Qualitative comparison on RealEstate10K with \cite{Tucker_2020_CVPR}. Our MINE generates much more photo-realestic outputs than \cite{Tucker_2020_CVPR}, it also inpaints the dis-occluded area much better (see hightlighted area in Row 3).} 
    \label{fig_realestate_vis}
\vspace{-4pt}
\end{figure*} 
\begin{figure*}[t!] \centering
\begin{tabular}{ccccc}
     Input & GT & MPI \cite{Tucker_2020_CVPR} & MINE ($N = 64$) \\ 
     \includegraphics[width=0.20\textwidth]{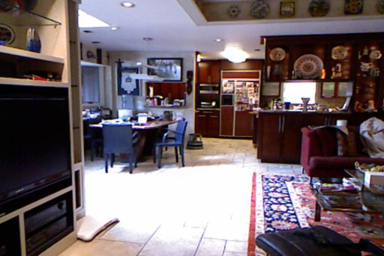} & \includegraphics[width=0.20\textwidth]{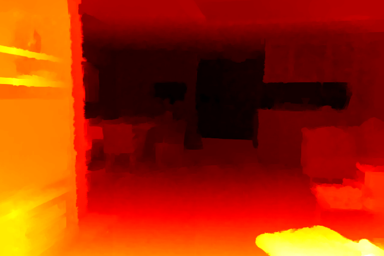} & \includegraphics[width=0.20\textwidth]{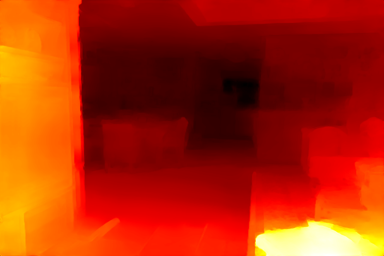} & \includegraphics[width=0.20\textwidth]{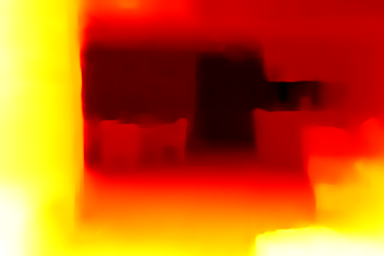} \\
     \includegraphics[width=0.20\textwidth]{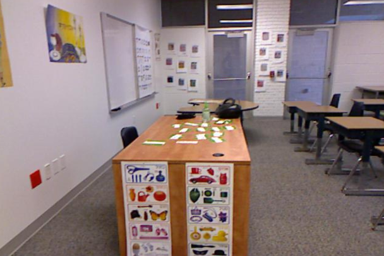} & \includegraphics[width=0.20\textwidth]{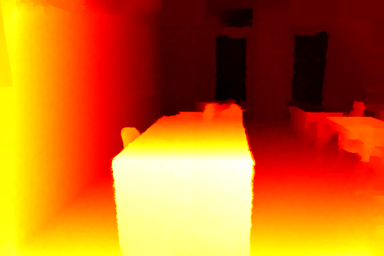} & \includegraphics[width=0.20\textwidth]{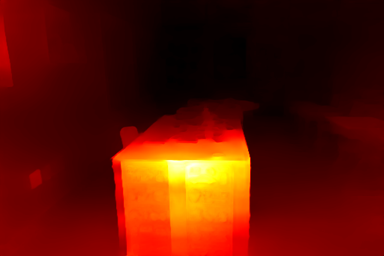} & \includegraphics[width=0.20\textwidth]{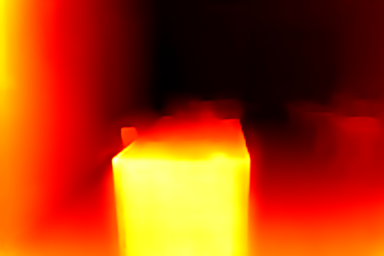} \\
     \includegraphics[width=0.20\textwidth]{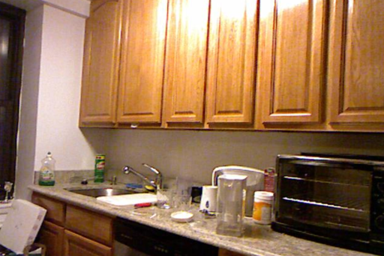} & \includegraphics[width=0.20\textwidth]{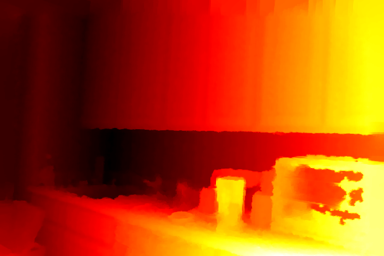} & \includegraphics[width=0.20\textwidth]{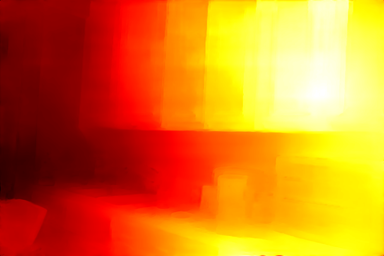} & \includegraphics[width=0.20\textwidth]{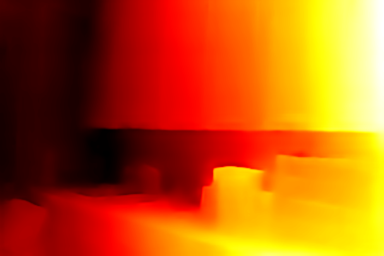} \\
\end{tabular}
    \caption{Qualitative comparison for disparity maps.} \vspace{-8pt}
    \label{fig_depth_vis}
\end{figure*} 

RealEstate10K \cite{zhou2018stereo} is a large-scale dataset of walk-through videos that contains both indoor and outdoor scenes. The dataset consists of $>70,000$ video sequences, which are pre-split into a training set and a test set. Each sequence contains the video frames and their corresponding camera intrinsic and extrinsic. To obtain the sparse 3D point clouds for scale-invariant learning and sparse depth supervision, we use COLMAP \cite{schoenberger2016sfm, schoenberger2016mvs} to perform SfM on each video sequence. For testing, we randomly sample 600 sequences from the official test split, and then we draw 5 frames from each sequence as the source frames. This gives us 3,000 source frames in total. Following \cite{Tucker_2020_CVPR}, we choose the target frames to be 5 or 10 frames apart for each reference frame. Additionally, we randomly sample another target frame from within 30 frames apart form the reference frame to create a more challenging setting.

In the RealEstate10K experiments, $N$ is set to 32 or 64, $\lambda_{smooth}=0.01$, and $\lambda_{d}$, $\lambda_{rgb}$, $\lambda_{SSIM}$ are set to 1.0. The disparity range is $[1.0, 0.001]$ for both the 32- and 64-plane models. The input resolution is set to $384\times256$. We train our models on 48 NVIDIA V100 SXM2 GPUs. We use the Adam Optimizer \cite{DBLP:journals/corr/KingmaB14} with an initial learning rate of 0.0002 for the encoder, and 0.001 for the decoder, we train the models for a total of 1,000,000 steps and the learning rate is decayed once at step 600,000. In training, we randomly sample the source frames, and sample the target frame within 30 frames apart from the source frames.
For fair comparisons, we run the MPI \cite{Tucker_2020_CVPR} open-sourced models 
on our test set, and obtained similar scores as the original paper. As shown in Tab. \ref{table:realestate_results}, it is clear that MINE outperforms both \cite{Wiles_2020_CVPR, Tucker_2020_CVPR} in all 3 criteria by a large margin. We show again that increasing $N$ from 32 to 64 gives better results, which is consistent with the ablation studies in KITTI. In Fig.~\ref{fig_realestate_vis}, we show qualitative comparison with \cite{Tucker_2020_CVPR}. MINE generates sharp and realistic target images, while \cite{Tucker_2020_CVPR} produces unpleasant artefacts and distortions. In particular, we highlight our inpainting capacity with a yellow bounding box.
\subsection{Depth Estimation on iBims-1 and NYU-V2}
We evaluate our depth estimation on the iBims-1 \cite{Koch18:ECS} and NYU-Depth V2 \cite{Silberman:ECCV12} benchmarks. Both benchmarks contain indoor scenes with dense ground truth depths. We use the model trained with RealEstate10K to synthesize the disparity maps and measure our depth estimation performance. Following \cite{Tucker_2020_CVPR,Niklaus_TOG_2019}, to solve the scale ambiguity of depth from single image, we scale and bias the depth predictions to minimize the $L2$ depth error before evaluation.
We compare with Depth in the Wild \cite{NIPS2016_0deb1c54}, MegaDepth \cite{MegaDepthLi18}, 3DKenBurns \cite{Niklaus_TOG_2019} and MiDaS \cite{Ranftl2020}, which are state-of-the-art systems trained with ground truth depth supervision. We also compare with~\cite{Tucker_2020_CVPR} which uses the same RGB video supervision as ours. Tab. \ref{table_depth_estimation} shows the quantitative results. Notably, even though MINE does not use any ground truth depth supervision in training, we achieve comparable performance as 3DKenBurns \cite{Niklaus_TOG_2019}, and significantly outperforms the other methods by a large margin. We further show qualitative comparison with \cite{Tucker_2020_CVPR} in Fig.~\ref{fig_depth_vis}. We find that MPI is easily biased towards image textures and thus producing unpleasant artefacts in the disparity maps, while MINE is able to generate smooth and more accurate disparity maps for images with texture-rich surfaces.
\subsection{View Synthesis on Flowers Light Fields} 
The Flowers light fields dataset \cite{Srinivasan_2017_ICCV} consists of 3,343 light fields captured with the Lytro Illum camera. Each light field has $14\times14$ angular samples and $376\time541$ spatial samples. Following \cite{Srinivasan_2017_ICCV} and \cite{Tucker_2020_CVPR}, we use the central $8\times8$ grids in our experiments to avoid using the angular samples outside the aperture. In testing, the centre image of the $8\times8$ grid is the source image and the four corner ones are the target images.
For comparison, we obtain the training and testing splits from \cite{Tucker_2020_CVPR}. Following their experiment settings, we randomly adjust the gamma from $[0.3, 0.7]$ in training and fixing it to 0.5 during testing. Since the scale is constant in this dataset, we set the scale factor to 1 following \cite{Tucker_2020_CVPR}. As shown in Tab.~\ref{tabel_flowers_results}, MINE improves upon \cite{Srinivasan_2017_ICCV} and \cite{Tucker_2020_CVPR}. As expected, increasing $N$ brings consistent improvements.

\section{Conclusion}  \label{sec_conclusion}
We propose MINE that is a continuous depth generalization of MPI by introducing NeRF. Given a single image, we jointly do a dense reconstruction of the camera frustum and inpainting of the occluded contents. We render our reconstructed frustum into novel view RGB images and depth maps with differentiable rendering.  Extensive experiments show that our method significantly outperforms existing state-of-the-art single-image view synthesis methods, and achieves near state-of-the-art performance on depth estimation without dense ground truth depth supervision.

\paragraph{Acknowledgment} \label{sec_ack}
This work is supported in part by the Singapore MOE Tier 2 grant MOE-T2EP20120-0011.

\begin{table}[t]
\small 
\centering
\begin{tabular}{c|ccc}
\hline 
Method                                            & LPIPS$\downarrow$  & SSIM$\uparrow$  & PSNR$\uparrow$ \\
\hline
Srinivasan et al, full \cite{Srinivasan_2017_ICCV} & -      & 0.822 & 28.1 \\
Tucker et al. \cite{Tucker_2020_CVPR}              & -      & 0.851 & 30.1 \\
MINE ($N = 32$)                                    & 0.1603 & 0.868 & 30.2 \\
MINE ($N = 64$)                                    & \textbf{0.1559} & \textbf{0.872} & \textbf{30.3} \\
\hline
\end{tabular}
\caption{View Synthesis on flowers light fields.}\label{tabel_flowers_results}
\vspace{-8pt}
\end{table}

%


{\small
\bibliographystyle{ieee_fullname}
\bibliography{ICCV2021_REF}
}




\clearpage
\title{\vspace{-12pt}MINE: Towards Continuous Depth MPI with NeRF for Novel View Synthesis - Supplementary Materials\vspace{-12pt}}
\maketitle
\appendix

\section{MINE vs. pixelNeRF and GRF}

There are two recent works: pixelNeRF~\cite{yu2020pixelnerf} and GRF~\cite{grf2020} that condition NeRF \cite{mildenhall2020nerf} on input image(s). pixelNeRF \cite{yu2020pixelnerf} first extracts a feature map from a given input image.
A feature vector at each query position $x$ and viewing direction $d$ is subsequently sampled from the feature map via projection and bilinear interpolation. The sampled feature vector then serves as an additional input to the MLP along with $x$ and $d$ to predict the RGB-$\sigma$ values. The rendering process is the same as NeRF. GRF \cite{grf2020} follows the same principles, but it assumes multiple views of a scene are available at test time. 

Our MINE is different from pixelNeRF and GRF in two aspects:

\begin{itemize}
    \item MINE directly models the frustum of the source camera, while both pixelNeRF and GRF model the entire 3D space. 
    \item MINE reconstructs the frustum of the source camera per plane, while pixelNeRF and GRF reconstruct the entire 3D space per ray.
\end{itemize}
A direct consequent of these differences is that our MINE is significantly more efficient. Both pixelNeRF and GRF render the output image pixel by pixel, and therefore the number of forward passes required is proportional to the spatial resolution of the output, the number of points along each ray, and the number of target views to render. 
On the contrary, since our MINE reconstructs the entire frustum of the source camera per plane, we only require $N_\text{{planes}}$ forward passes of the fully-convolutional decoder to obtain the representation. Furthermore, the rendering for each novel view only requires an additional homography warping step. 

More concretely, let us denote the output resolution as $H \times W$. 
We further denote the number of points along each ray from pixelNeRF and GRF as $N_{\text{points}}$, and the number of planes from our MINE as $N_{\text{planes}}$. 
The number of network forward passes $P$ required for these methods are:

\begin{equation}
\begin{split}
        & P_\text{{pixelNeRF}} = 1 + N_\text{{targets}} \times N_\text{{points}} \times H \times W, \\
   & P_\text{{GRF}} = N_\text{{views}} + N_\text{{targets}} \times N_\text{{points}} \times H \times W, \\
   & P_\text{{MINE}} = 1 + N_\text{{planes}}, \\
\end{split}
\end{equation}
where $N_\text{{targets}}$ denotes the number of novel views. 

All three methods listed above utilize the encoder-decoder structure to condition on the input image(s). pixelNeRF and our MINE takes single image as input, and then the encoder is forwarded only $1$ time. In contrast, GRF takes multiple images as input, and thus requires $N_{\text{views}}$ encoder inferences.

Note that for pixelNeRF and GRF, $N_\text{{points}} \times H \times W$ times of decoder (MLP) inferences are required for \textbf{each} target view. On the other hand, our MINE reconstructs the frustum using $N_\text{{planes}}$ decoder (Fully Convolutional Network) inferences. After the reconstruction, only homography warping is required to render into $any$ target view. Consequently, the complexity of our MINE is independent of $N_{\text{targets}}$, while the complexity of pixelNeRF and GRF is proportional to $N_{\text{targets}}$.

Also note that our method does not take viewing direction as inputs, but we argue that the viewing directions can be easily integrated into our framework by concatenating the per-ray viewing directions with the output feature maps at the target view (after warping), and then using a lightweight fully convolutional network to predict the view-dependent radiance. This only adds $N_{\text{views}}$ network inferences in total, which is still significantly faster than pixelNeRF and GRF.


The efficiency of our MINE also allows for more flexible training strategies. Since our MINE renders the full target image and disparity map in training time, it is possible to impose dense supervision signals, e.g. SSIM \cite{1284395} and edge-aware smoothness loss \cite{monodepth2, monodepth17, Tucker_2020_CVPR}. We argue that these dense supervision signals are helpful for generalizing to real-world large-scale datasets, as verified empirically by our extensive experiments. Due to their inefficiency, it is infeasible for NeRF-like methods to render the full image at training time.

Lastly, neither pixelNeRF nor GRF presents experiments with large scale real-worlds data. Our MINE is verified with well-known datasets like KITTI, NYU-V2, RealEstate10k, etc.

\section{Additional Implementation Details} \label{sec_impl_details}

\paragraph{Network architecture.} Our encoder is a standard ResNet50 \cite{resnet2016}, we take the outputs of [conv1, layer1, layer2, layer3, layer4] as the final output of the encoder. We give a complete description of our decoder architecture in Table \ref{table_decoder_arch}. The decoder is the same as the depth decoder in \cite{monodepth2}, except that we add two additional downsampling blocks and two upsampling blocks to increase the receptive fields of the network, and the output of the network is a 4-channel RGB-$\sigma$ image. The RGB output is produced by a Sigmoid layer, and $\sigma$ is produced by taking the absolute value of the last channel of the output. We adopt the multi-scale training strategy in \cite{monodepth2} with the exception that $\mathcal{L}_{L1}$ and $\mathcal{L}_{SSIM}$ are only applied on output1 and $\mathcal{L}_{smooth}$ is applied on all [output1, output2, output3, output4]. Note that our method is not restricted to specific network architecture.

\begin{table*}[]
\centering
\resizebox{\textwidth}{!}
{
\small
\begin{tabular}{ c | c | c | c | c | c}
\hline
layer       & k & in-channels & out-channels & input                          & activation                               \\
\hline
downconv1 & 1 & 2048        & 512          & encoder\_layer4                         & ELU \cite{DBLP:journals/corr/ClevertUH15}                                      \\
downconv2 & 3 & 512         & 256          & downconv1                    & ELU                                      \\
upconv1\_extra   & 3 & 256         & 256          & downconv2                    & ELU                                      \\
upconv2\_extra   & 1 & 256         & 2048         & upconv1\_extra                      & ELU                                      \\
\hline
upconv5     & 3 & 2048 + 21   & 256          & cat(upconv2\_extra, disparity\_encoding) & ELU                                      \\
iconv5      & 3 & 256 + 1024 + 21         & 256          & cat(upconv5, encoder\_layer3, disparity\_encoding)               & ELU                                      \\
upconv4     & 3 & 256         & 128          & iconv5                         & ELU                                      \\
iconv4      & 3 & 128 + 512 + 21        & 128          & cat(upconv4, encoder\_layer2, disparity\_encoding)               & ELU                                      \\
output4     & 3 & 128         & 4            & iconv4                         & Sigmoid (for RGB) and abs (for $\sigma$) \\
upconv3     & 3 & 128         & 64           & iconv4                         & ELU                                      \\
iconv3      & 3 & 64 + 256 + 21          & 64           & cat(upconv3, encoder\_layer1, disparity\_encoding)               & ELU                                      \\
output3     & 3 & 64          & 4            & iconv3                         & Sigmoid (for RGB) and abs (for $\sigma$) \\
upconv2     & 3 & 64          & 32           & iconv3                         & ELU                                      \\
iconv2      & 3 & 32 + 64 + 21         & 32           & cat(upconv2, encoder\_conv1, disparity\_encoding)               & ELU                                      \\
output2     & 3 & 32          & 4            & iconv2                         & Sigmoid (for RGB) and abs (for $\sigma$) \\
upconv1     & 3 & 32          & 16           & iconv2                         & ELU                                      \\
iconv1      & 3 & 16          & 16           & upconv1                        & ELU                                      \\
output1     & 3 & 16          & 4            & iconv1                         & Sigmoid (for RGB) and abs (for $\sigma$) \\
\hline
\end{tabular}
}
\caption{Network architecture for our depth decoder. All upconv blocks consist of a convolution layer, a batch normalization layer and an activation layer as specified in the table, followed by a $2\times$ nearest neighbour upsampling. The downconv blocks consist of a max pooling layer of stride 2, a convolution layer followed by an activation layer.} \label{table_decoder_arch}
\end{table*}

\paragraph{Pre-processing for Flowers Light Fields.} For the Flower Light Fields  dataset \cite{Srinivasan_2017_ICCV}, we set the disparity range to be $[3.0, 0.03]$. Since this dataset was taken with the Lytro Illum camera, there is a shift in the principle point between different views. In this dataset, all light fields are taken with the same camera, but the shifts in the principle points could vary across different scenes. Since there is no metadata to indicate the amount of the shifts, we follow \cite{Tucker_2020_CVPR} and make it constant. Specifically, we set the camera intrinsics as follows: 
\begin{equation}
\begin{split}
    & f_x = 0.868056, \quad f_y = 1.250000, \\
    & c_{x_{ij}} = 0.5 + 0.002667 * i, \\
    & c_{y_{ij}} = 0.5 + 0.002667 * j, 
\end{split}
\end{equation}
where $[i, j]$ is the index in the extracted $8\times8$ grid and $[0, 0]$ denotes the top left view. Since this dataset was captured with a light field camera, in addition to the principle point shift, there is a translation between different views and there is no rotation. In training and testing, same as \cite{Tucker_2020_CVPR}, we set the distance between adjacent grids to be $0.00128$.

\section{Additional Qualitative Results} \label{sec_qualitative_results}

\begin{figure*}[t!] \centering
\setlength{\tabcolsep}{1pt}
\begin{tabular}{ccccc}
     Input & Target GT & Input Disparity & Synthesised Target \\ 
     \includegraphics[width=0.25\textwidth]{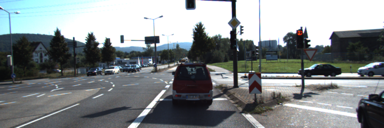} & \includegraphics[width=0.25\textwidth]{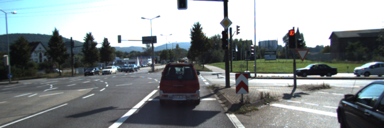} & \includegraphics[width=0.25\textwidth]{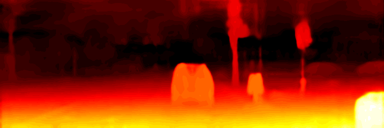} & \includegraphics[width=0.25\textwidth]{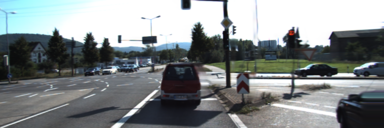} \\
     \includegraphics[width=0.25\textwidth]{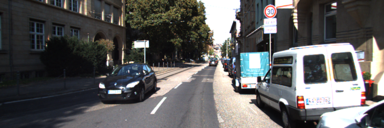} & \includegraphics[width=0.25\textwidth]{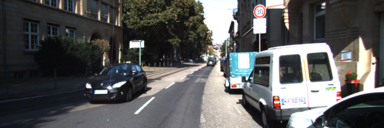} & \includegraphics[width=0.25\textwidth]{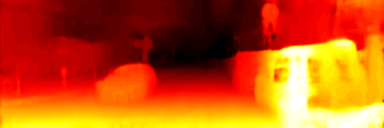} & \includegraphics[width=0.25\textwidth]{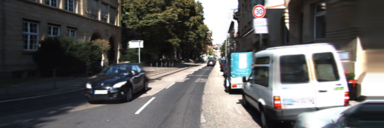} \\
     \includegraphics[width=0.25\textwidth]{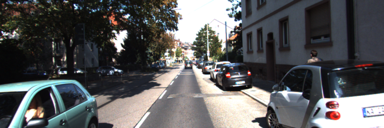} & \includegraphics[width=0.25\textwidth]{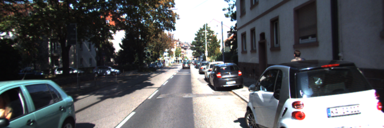} & \includegraphics[width=0.25\textwidth]{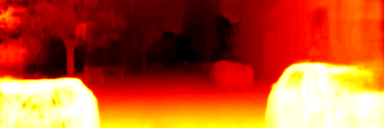} & \includegraphics[width=0.25\textwidth]{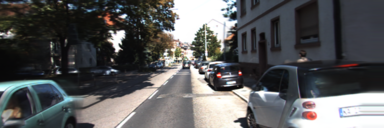} \\
     \includegraphics[width=0.25\textwidth]{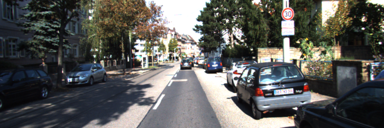} & \includegraphics[width=0.25\textwidth]{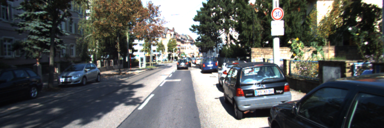} & \includegraphics[width=0.25\textwidth]{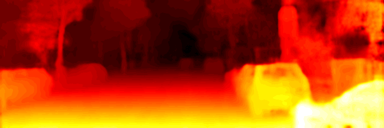} & \includegraphics[width=0.25\textwidth]{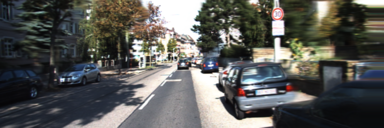} \\
     \includegraphics[width=0.25\textwidth]{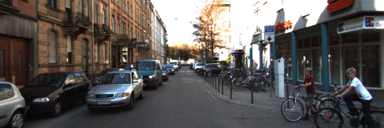} & \includegraphics[width=0.25\textwidth]{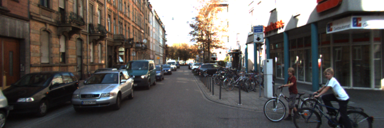} & \includegraphics[width=0.25\textwidth]{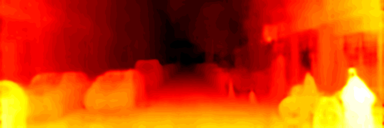} & \includegraphics[width=0.25\textwidth]{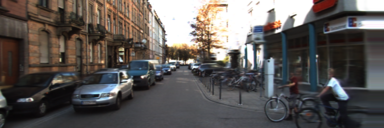} \\
     \includegraphics[width=0.25\textwidth]{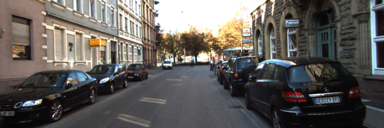} & \includegraphics[width=0.25\textwidth]{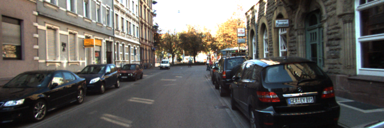} & \includegraphics[width=0.25\textwidth]{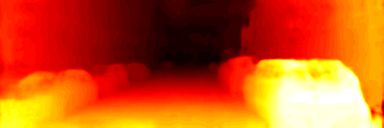} & \includegraphics[width=0.25\textwidth]{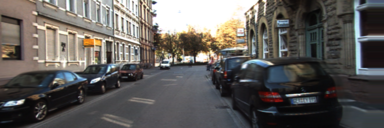} \\
     \includegraphics[width=0.25\textwidth]{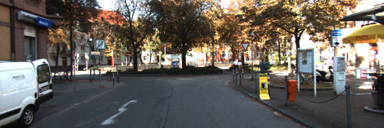} & \includegraphics[width=0.25\textwidth]{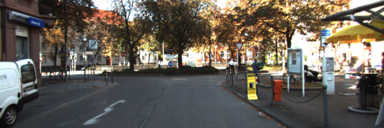} & \includegraphics[width=0.25\textwidth]{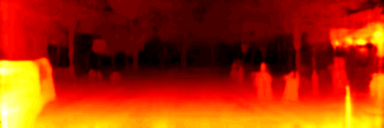} & \includegraphics[width=0.25\textwidth]{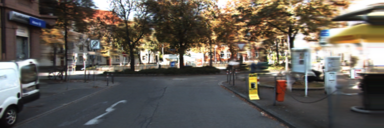}
\end{tabular}
    \caption{Qualitative results for KITTI.} 
    \label{fig_sm_kitti_vis}
\end{figure*}

\begin{figure*}[t!] \small \centering
\begin{tabular}{ccccc}
     Input & Target GT & Input Disparity & Synthesised Target \\ 
     \includegraphics[width=0.22\textwidth]{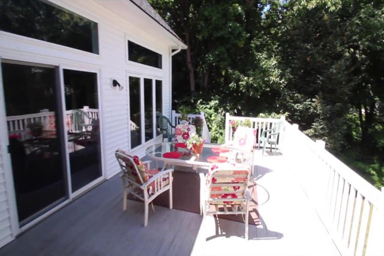} & \includegraphics[width=0.22\textwidth]{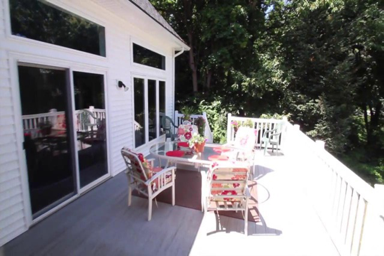} & \includegraphics[width=0.22\textwidth]{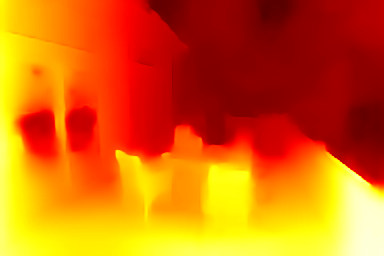} & \includegraphics[width=0.22\textwidth]{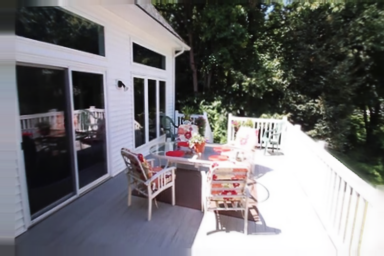} \\
     \includegraphics[width=0.22\textwidth]{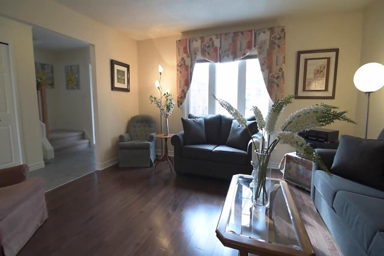} & \includegraphics[width=0.22\textwidth]{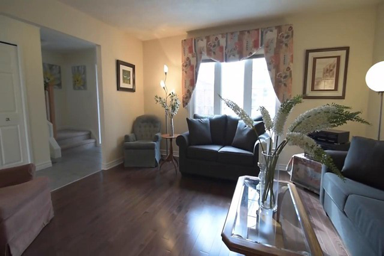} & \includegraphics[width=0.22\textwidth]{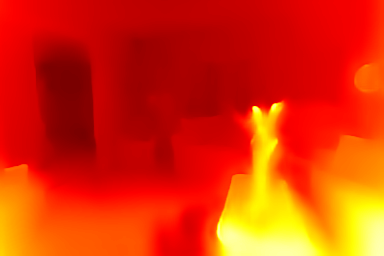} & \includegraphics[width=0.22\textwidth]{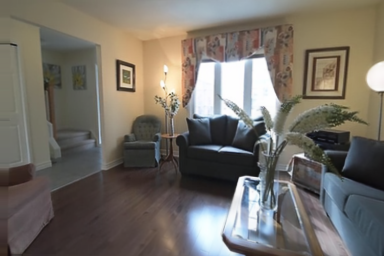} \\
     \includegraphics[width=0.22\textwidth]{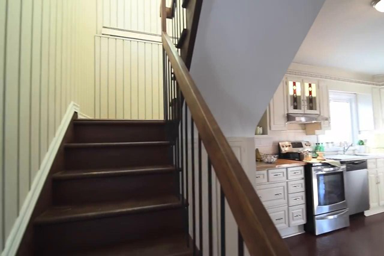} & \includegraphics[width=0.22\textwidth]{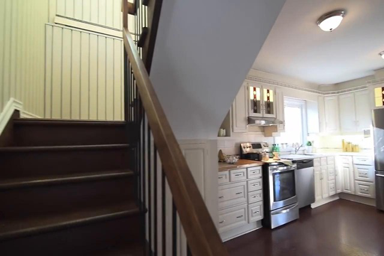} & \includegraphics[width=0.22\textwidth]{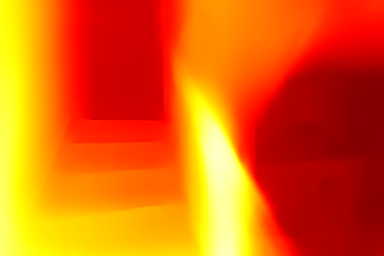} & \includegraphics[width=0.22\textwidth]{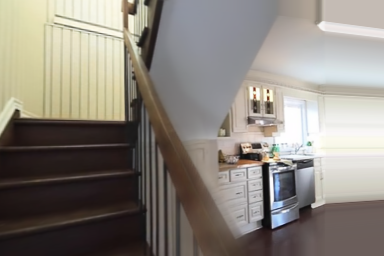} \\
     \includegraphics[width=0.22\textwidth]{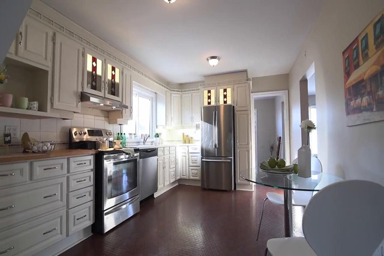} & \includegraphics[width=0.22\textwidth]{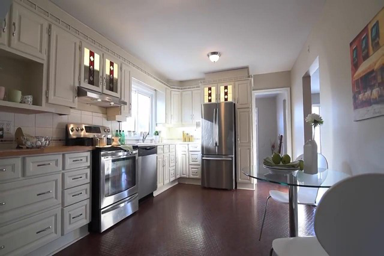} & \includegraphics[width=0.22\textwidth]{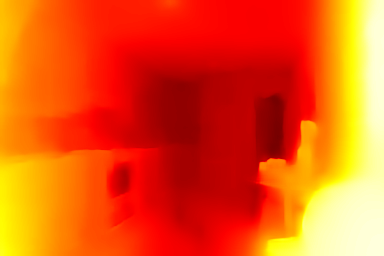} & \includegraphics[width=0.22\textwidth]{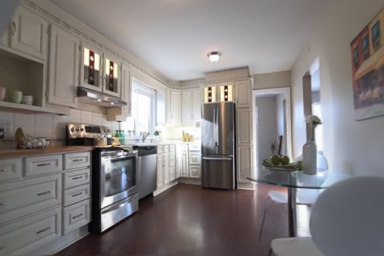} \\
     \includegraphics[width=0.22\textwidth]{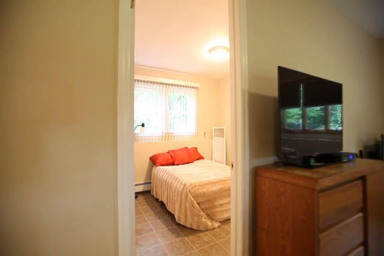} & \includegraphics[width=0.22\textwidth]{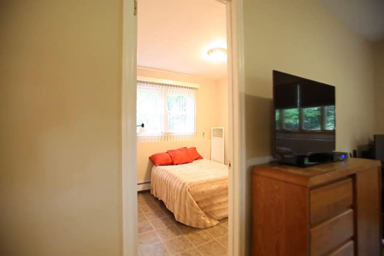} & \includegraphics[width=0.22\textwidth]{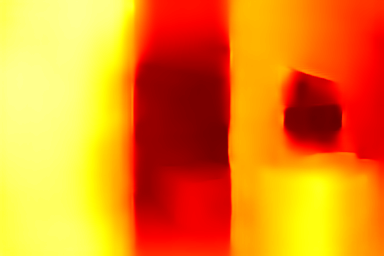} & \includegraphics[width=0.22\textwidth]{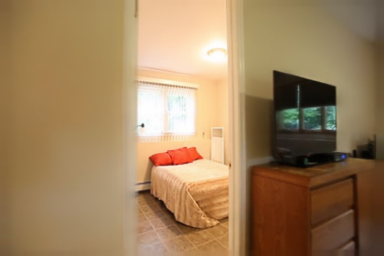} \\
\end{tabular}
    \caption{Qualitative results for RealEstate10K.} 
    \label{fig_sm_realestate_vis}
\end{figure*}

\begin{figure*}[t!] \small \centering
\begin{tabular}{ccccc}
     Input & Target GT & Input Disparity & Synthesised Target \\ 
     \includegraphics[width=0.22\textwidth]{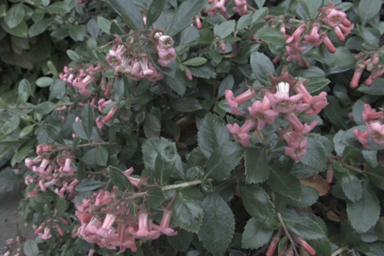} & \includegraphics[width=0.22\textwidth]{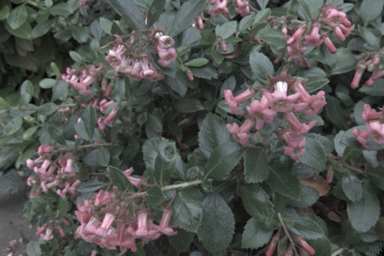} & \includegraphics[width=0.22\textwidth]{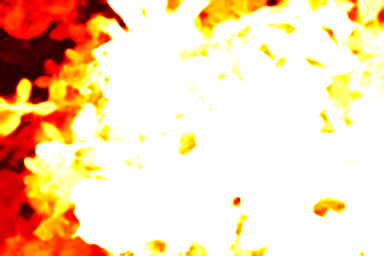} & \includegraphics[width=0.22\textwidth]{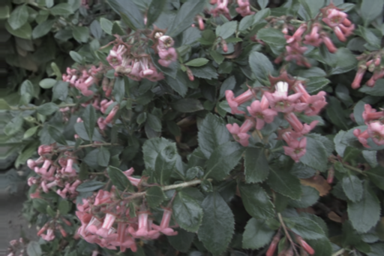} \\
     \includegraphics[width=0.22\textwidth]{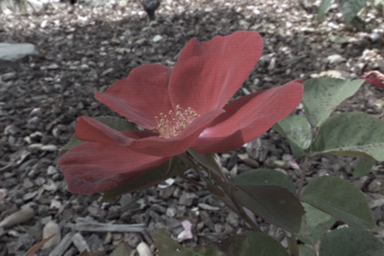} & \includegraphics[width=0.22\textwidth]{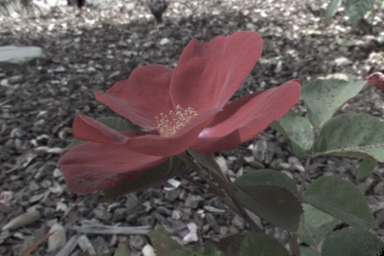} & \includegraphics[width=0.22\textwidth]{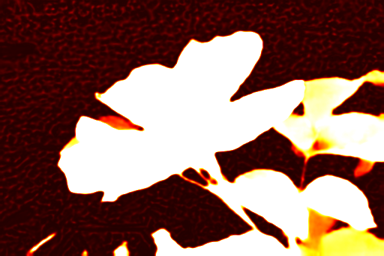} & \includegraphics[width=0.22\textwidth]{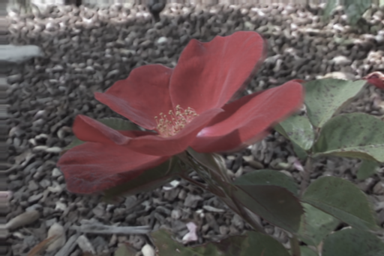} \\
     \includegraphics[width=0.22\textwidth]{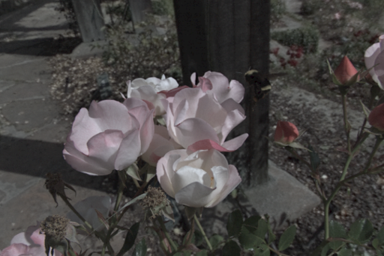} & \includegraphics[width=0.22\textwidth]{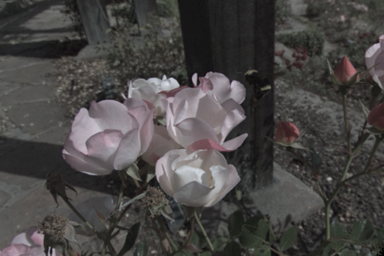} & \includegraphics[width=0.22\textwidth]{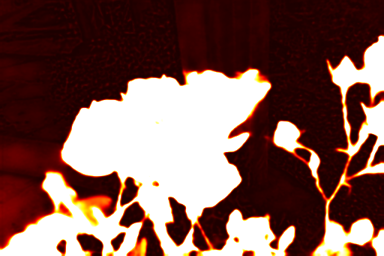} & \includegraphics[width=0.22\textwidth]{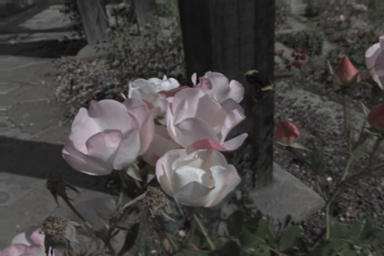} \\
     \includegraphics[width=0.22\textwidth]{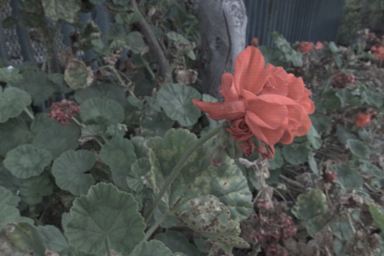} & \includegraphics[width=0.22\textwidth]{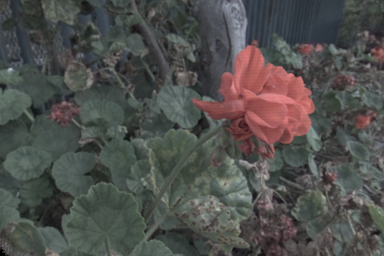} & \includegraphics[width=0.22\textwidth]{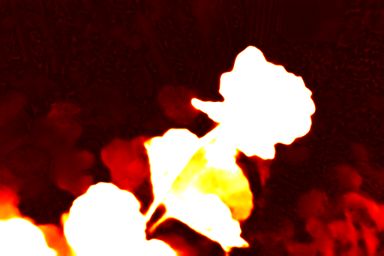} & \includegraphics[width=0.22\textwidth]{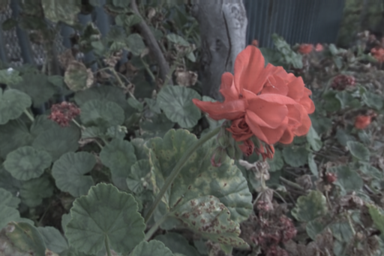} \\
     \includegraphics[width=0.22\textwidth]{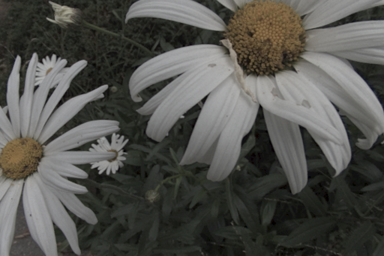} & \includegraphics[width=0.22\textwidth]{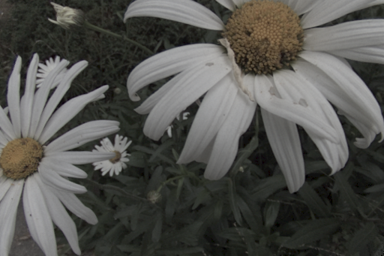} & \includegraphics[width=0.22\textwidth]{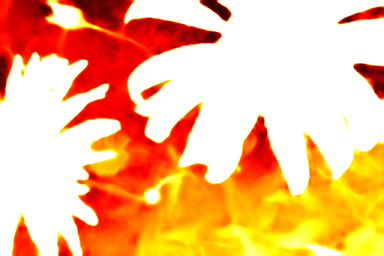} & \includegraphics[width=0.22\textwidth]{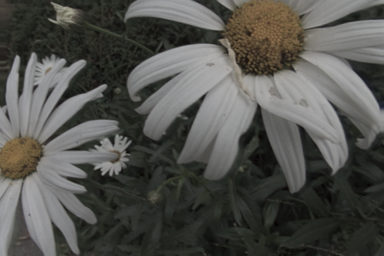} \\
     \includegraphics[width=0.22\textwidth]{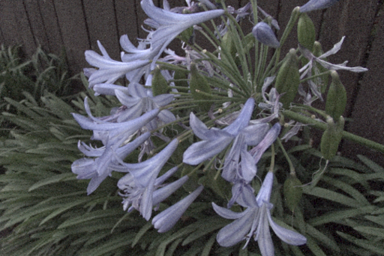} & \includegraphics[width=0.22\textwidth]{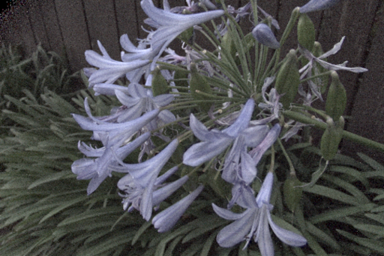} & \includegraphics[width=0.22\textwidth]{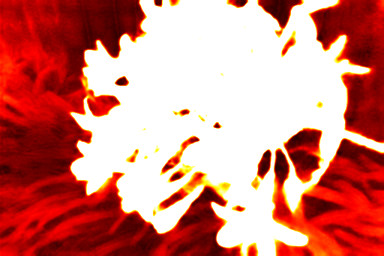} & \includegraphics[width=0.22\textwidth]{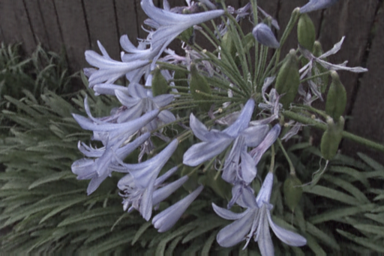}
\end{tabular}
    \caption{Qualitative results for Flowers Light Fields.} 
    \label{fig_sm_flowers_vis}
\end{figure*}

\paragraph{Supplementary image results.} We include additional qualitative results for KITTI (Figure \ref{fig_sm_kitti_vis}), RealEstate10K \cite{zhou2018stereo} (Figure \ref{fig_sm_realestate_vis}) and Flowers Light Fields (Figure \ref{fig_sm_flowers_vis}). Our method generalizes well to a wide range of real-world scenes, including outdoor and indoor scenes, and flowers with complex geometry. All scenes are unseen in training.

\vspace{-10pt}
\paragraph{Supplementary video results.} We also include supplementary videos results (uploaded separately) for the RealEstate10K, KITTI and iBims-1 \cite{Koch18:ECS} datasets, covering both outdoor scenes and indoor scenes with complex geometries and textures. For each scene, we include both the RGB videos and the videos of disparity maps. Given a single image as input, we generate the video by rendering into multiple novel views. All scenes are unseen during training. We demonstrate that even under large camera motion, our MINE is still able to generate temporally consistent realistic images, and smooth and accurate disparity maps.

\clearpage

\end{document}